\documentclass[conference]{IEEEtran}
\usepackage{cite}
\usepackage{amsmath,amssymb,amsfonts}
\usepackage{algorithmic}
\usepackage{graphicx}
\usepackage{textcomp}
\usepackage{xcolor}

\usepackage[ruled,vlined]{algorithm2e}
\usepackage{booktabs}
\usepackage{multirow}
\usepackage{lscape}
\usepackage{pifont}
\usepackage[pagebackref,breaklinks,colorlinks]{hyperref}

\usepackage[normalem]{ulem}
\usepackage{subcaption}
\usepackage[table]{colortbl}

\def\BibTeX{{\rm B\kern-.05em{\sc i\kern-.025em b}\kern-.08em
    T\kern-.1667em\lower.7ex\hbox{E}\kern-.125emX}}

\usepackage[textsize=small]{todonotes}

\usepackage{authblk}

\begin{document}

\title{
FLoRA: Enhancing Vision-Language Models with Parameter-Efficient Federated Learning}

\newcommand{\proj}{\textsc{FLoRA}\xspace}

\newcommand{\cmark}{\ding{51}}%
\newcommand{\xmark}{\ding{55}}%

\author[1]{Duy Phuong Nguyen}
\author[2]{J. Pablo Mu{\~{n}}oz}
\author[1]{Ali Jannesari}
\affil[1]{Department of Computer Science, Iowa State University, Ames, IA, USA}
\affil[2]{Intel Corporation, Santa Clara, CA, USA}
\affil[1]{\textit {\{dphuong, jannesari\}@iastate.edu}}
\affil[2]{\textit {pablo.munoz@intel.com}}

\maketitle

\begin{abstract}

In the rapidly evolving field of artificial intelligence, multimodal models, e.g., integrating vision and language into visual-language models (VLMs), have become pivotal for many applications, ranging from image captioning to multimodal search engines. Among these models, the Contrastive Language–Image Pre-training (CLIP) model has demonstrated remarkable performance in understanding and generating nuanced relationships between text and images. However, the conventional training of such models often requires centralized aggregation of vast datasets, posing significant privacy and data governance challenges. To address these concerns, this paper proposes a novel approach that leverages Federated Learning and parameter-efficient adapters, i.e., Low-Rank Adaptation (LoRA), to train VLMs. This methodology preserves data privacy by training models across decentralized data sources and ensures model adaptability and efficiency through LoRA's parameter-efficient fine-tuning. Our approach accelerates training time by up to $34.72 \times$, and requires $2.47 \times$ less memory usage than full fine-tuning. 



\end{abstract}


\section{Introduction}

In recent years, the fields of computer vision and natural language processing have witnessed remarkable advancements fueled by the power of deep learning models~\cite{devlin2019bert, radford2019gpt2, brown2020gpt3}. Vision-language models (VLMs)~\cite{radford2021clip}, in particular, have gained significant attention due to their ability to bridge the gap between visual and textual information, enabling a wide range of applications such as image captioning, visual question answering, and text-to-image synthesis.
However, the success of VLMs heavily relies on the availability of vast amounts of labeled data for training. Collecting and annotating large-scale datasets can be time-consuming, resource-intensive, and potentially privacy-sensitive. Additionally, centralizing all data in a single location for training poses challenges such as data privacy, communication costs, and scalability.
Federated learning~(FL) has emerged as a promising approach to address these limitations. Federated learning allows training models across a distributed network of devices, where each device holds its data and collaboratively learns a shared model while keeping the data locally. This decentralized approach respects data privacy and reduces the communication overhead in transmitting raw data to a centralized server.

This paper delves into the domain of FL with a particular focus on fine-tuning VLMs. By adapting the fine-tuning techniques to a federated setting, we seek to harness the dual strengths of vision-language pre-training and distributed learning. Our method utilizes the LoRA (Low-Rank Adaptation)~\cite{hu2022lora} technique to fine-tune the CLIP model, aiming to update only a small subset of the model's parameters while keeping the bulk of the pre-trained weights intact.

\begin{figure}[ht]
\begin{center}


\centerline{\includegraphics[width=0.5\textwidth]{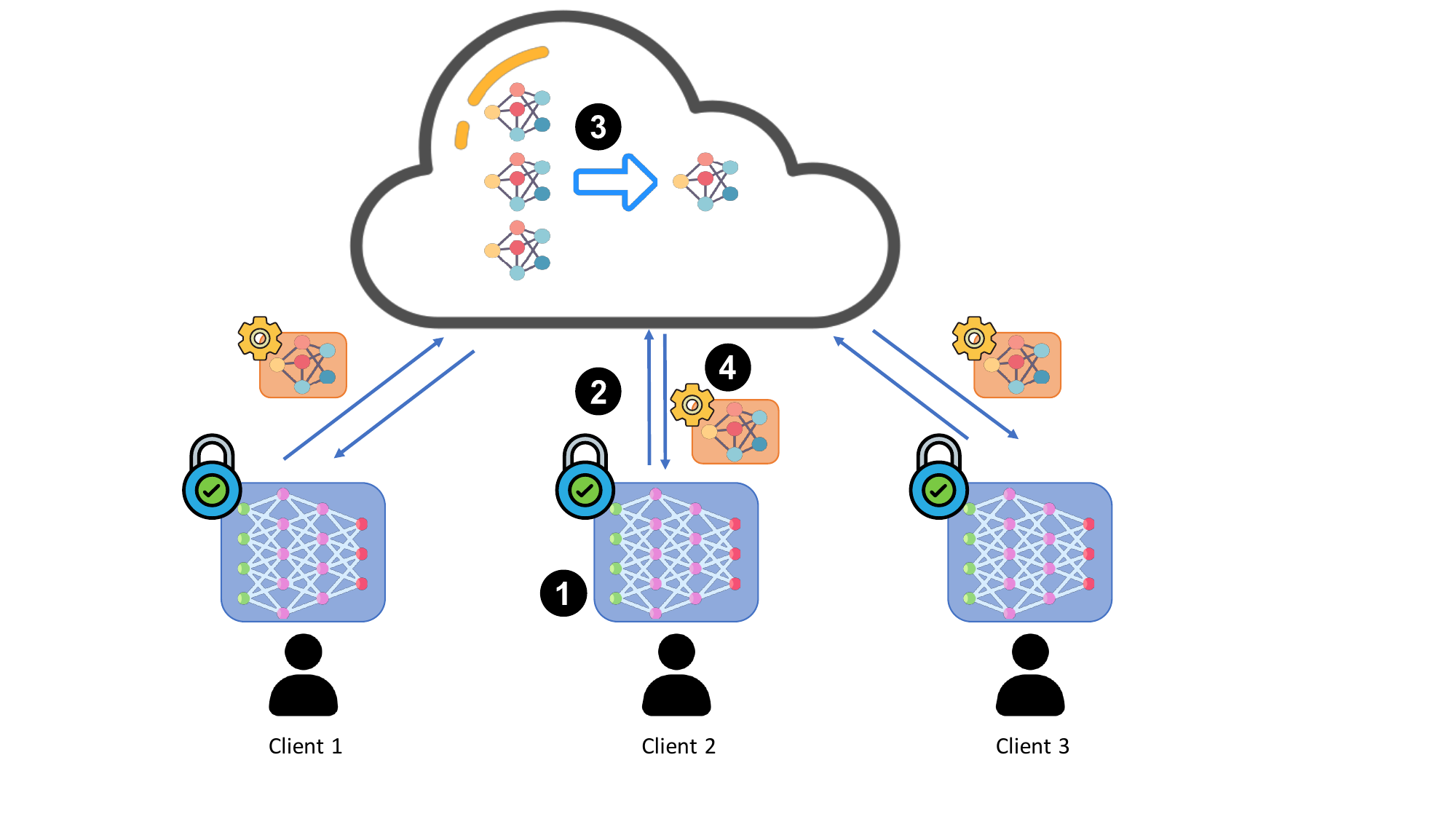}}

\caption{This schematic illustrates the critical steps in federated learning. In each round of federated learning, clients first downloads the transferred model from the server and independently train models on their own data~(step \ding{202}). Then, locally-trained models are then sent to a central server~(step~\ding{203}). After that, the server aggregates these models to form a average global model~(step~\ding{204}), which is subsequently redistributed to the clients for further training or inference~(step~\ding{205}).}

\vspace{-2.0em}

\label{fig:overview}
\end{center}
\end{figure}

Figure~\ref{fig:overview} illustrates the overarching process. We investigate efficient strategies for federated optimization, employing LoRA as a mechanism to perform fine-tuning that respects the limitations of data privacy and the necessity for reduced communication overhead. Our approach relies on the inherent alignment capabilities of the CLIP model, fine-tuning it to enhance its performance on tasks involving federated datasets. We investigate novel techniques for collaborative model training and federated optimization strategies that enable VLMs to learn from distributed data sources while preserving privacy and minimizing communication costs.

Through rigorous experimental evaluations, we strive to exhibit the effectiveness of federated fine-tuning, positing that our methodology achieves competitive performance on several benchmarks. The implications of our work extend to the advancement of FL techniques and open new pathways for the development of collaborative, privacy-aware VLMs.

\begin{figure*}[ht]
\begin{center}

\centerline{\includegraphics[width=0.7\textwidth]{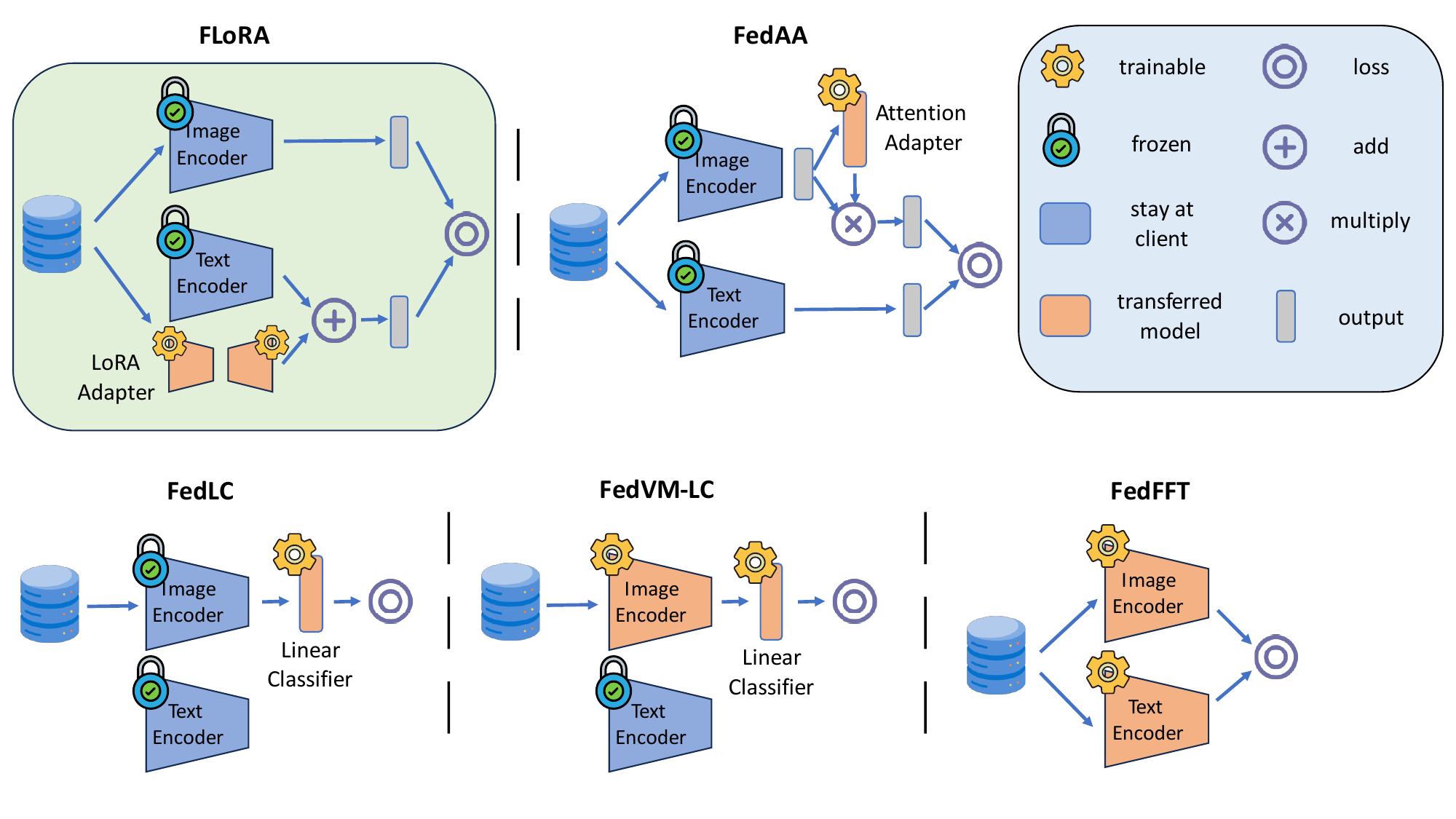}}

\caption{Different methods for local model update. The clients download the transferred model from the server, train locally with the local data, and upload it to the server. \textbf{\proj}: only transfer LoRA Adapter; \textbf{FedAA}: only transfer Attention Adapter; \textbf{FedLC}: only transfer Linear Classfier; \textbf{FedVM-LC}: only transfer Vision Model and Linear Classfier; \textbf{FedFFT}: transfer the whole CLIP model (full fine-tuning). Our method (\textbf{\proj}) is highlighted in the green box. The figure is best viewed in color.
}

\label{fig:method}
\end{center}

\vspace{-2.0em}

\end{figure*}

This paper embarks on an in-depth examination of federated fine-tuning learning's potential to revolutionize VLMs, addressing critical concerns such as data privacy, communication efficacy, and scalability in decentralized environments.

Our contributions are outlined as follows:
\begin{itemize}
    \item We pioneer the application of LoRA to VLMs within a FL framework, particularly focusing on the fine-tuning of the CLIP model's text encoder.
    
    \item 
    Our adaptation of the LoRA adapter into the CLIP model's text encoder projection layers not only refines the model's specificity to textual nuances but also results in up to a $4766 \times$ reduction in communication overhead during federated training—a significant stride towards practical FL deployment.

    \item 
    Our methodology, \proj, has been rigorously evaluated across datasets with varying characteristics, leading to performance gains of up to $30\%$ in accuracy metrics compared to traditional federated learning baselines, underscoring the robustness and versatility of our approach.

    \item 
    
    Our research includes an extensive ablation study, meticulously investigating the optimal integration points and methods for the LoRA adapter within the CLIP model. This investigation is pivotal in optimizing performance within the FL context, ensuring that our approach enhances efficiency and maximizes the efficacy of the model's learning capabilities.
\end{itemize}

In subsequent sections, we outline related work, delineate our methodology, present experimental results, and discuss the broader impact and future trajectory of federated fine-tuning for VLMs.


\begin{algorithm}[tb]
\caption{\proj Client Update}
\label{alg:local}

\SetKwInOut{KwIn}{Input}
\SetKwInOut{KwOut}{Output}

\KwIn{$N$: number of clients, $\rho$: client sample ratio, $\mathcal{L}$: Loss function, $\delta$: local learning rate, $\Theta^t$ previous global model}

\KwOut{local model $\Theta_1^{t+1}, \ldots, \Theta_N^{t+1}$}

$\Theta^t \rightarrow$ local models

$\mathcal{D}_i \rightarrow$ local data of client $i \in [1, N]$

Clients incorporate $\Theta^t$ into $\Theta_{\text{CLIP}}$

\For{\textsc{Client $i \in$ [$1, \rho \times N$]}}{

$\triangleright \textbf{ Local training}$

$\Theta_i^{t+1} \leftarrow \Theta_i^t - \delta \bigtriangledown [ \mathcal{L}(\Theta_i^t ; \mathcal{D}_i)] $ 

$\triangleright \textbf{ Uploading}$

$\text{Client $i$ sends $\Theta_i^{t+1}$ to the server}$
}

\end{algorithm}

\begin{algorithm}[tb]
\caption{\proj Server Update}
\label{alg:server}

\SetKwInOut{KwIn}{Input}
\SetKwInOut{KwOut}{Output}

\KwIn{$T$: number of communication rounds, $N$: number of clients, $\rho$: client sample ratio} 

\KwOut{Global model $\Theta^t$}

\If{t == 0}{

$\triangleright \textbf{ Initial stage}$

Initialize $\Theta^0$
}

Server receives $\Theta_i^{t}$ from client $i \in [1, N \times \rho]$

\For{\textsc{iteration $t = 1,\ldots, T$}} {

$\triangleright \textbf{ Averaging weights}$

$\Theta^{t} \leftarrow \sum_{i=1}^{\lfloor N \times \rho \rfloor}{\frac{|\mathcal{D}_i|}{ \sum_{j=1}^{\lfloor N \times \rho \rfloor} |\mathcal{D}_j|} \Theta_i^{t}}$

}

$\triangleright \textbf{ Downloading}$

Server sends $\Theta^{t}$ to all clients $i \in [1, N] $  

\end{algorithm}


\section{Related Work}

The synthesis of FL and VLMs has garnered substantial attention in recent times. This section encapsulates seminal contributions to these fields, laying the groundwork for our investigation into federated fine-tuning of VLMs.

\subsection{Federated Learning}
Federated learning, introduced by McMahan et al.~\cite{mcmahan2017fedavg}, has emerged as a powerful approach for training models collaboratively across distributed devices while preserving data privacy. Various studies have explored the application of FL in different domains, including computer vision and natural language processing. McMahan et al.~\cite{mcmahan2017fedavg} proposed federated averaging~(FedAvg), a communication-efficient algorithm for training deep learning models in a decentralized manner. Li et al.~\cite{li2020convergence_noniid} extended FL to handle large-scale datasets through hierarchical aggregation and compression techniques.


\subsection{Vision-Language Models}
Vision-language models serve as a conduit between the visual and linguistic realms, paving the way for multifaceted applications. Early innovations were predominantly in image captioning, as exemplified by the work of Vinyals et al.~\cite{vinyals2015pointer} and Xu et al.~\cite{xu2015show}. Subsequent models like ViLBERT~\cite{lu2019vilbert} and LXMERT~\cite{tan2019lxmert} capitalized on extensive pre-training to set new benchmarks across various VLMs.


\subsection{Contrastive Language Image Pre-training (CLIP)}
The CLIP model, unveiled by OpenAI~\cite{radford2021clip}, marked a paradigm shift by leveraging contrastive pre-training over a vast corpus of image-text pairs. It encodes images and text into a shared vector space, facilitating versatile zero-shot capabilities. This property has allowed CLIP to deliver impressive results across diverse datasets, even outperforming models fine-tuned on specific benchmarks such as ImageNet and its variants~\cite{russakovsky2015imagenet}.

\subsection{Federated Learning for Vision-Language Models}

Bridging FL with VLMs remains an emergent research domain. Notable are the work by Guo et al.~\cite{guo2023promptfl,guo2023pfedprompt}, which proposes a privacy-centric federated framework for image classification. Yet, the fusion of FL with the complexity of vision-language tasks necessitates further exploration. Qiu et al~\cite{qiu2024fedtpg} proposed FedTPG, a method that tailors pre-trained VLMs to FL by employing task-specific prompt generation. This technique enhances the relevance of the generated prompts to each client's data, thereby improving the model's performance on localized tasks. However, FedTPG's prompt generation process still relies on the exchange of prompt-related parameters, which could be optimized to reduce communication costs further. Another one by Lu et al.~\cite{lu2023fedclip} demonstrates the feasibility of adapting CLIP to federated settings, yet it leaves room for efficiency improvements, particularly in communication overhead and model adaptability. Our work extends these efforts, integrating FL with the fine-tuning of VLMs, eschewing traditional prompt learning for a focused adaptation using LoRA. The process is visualized in Figure~\ref{fig:overview}, marking a significant stride toward privacy-conscious, efficient, and scalable vision-language model training, especially in terms of reducing the bandwidth required for transmitting updates and ensuring the model remains flexible to the nuances of distributed datasets.

\subsubsection{Client-Side Training}
In our framework, client-side training involves local updates using a LoRA-adapted CLIP model, where Adam optimizer~\cite{kingma2015adam} aids in gradient-based optimization. This process, outlined in Algorithm~\ref{alg:server}, ensures that updates are focused and computationally frugal.

\subsubsection{Server Aggregation}
Our approach employs a server that orchestrates the aggregation of updates from clients, utilizing a weighted averaging mechanism akin to FedAvg~\cite{mcmahan2017fedavg}. This strategy balances the contributions across diverse datasets, promoting an egalitarian learning process. Details of this implementation are provided in Algorithm~\ref{alg:local}.

\begin{figure}[ht]
\begin{center}


\centerline{\includegraphics[width=0.4\textwidth]{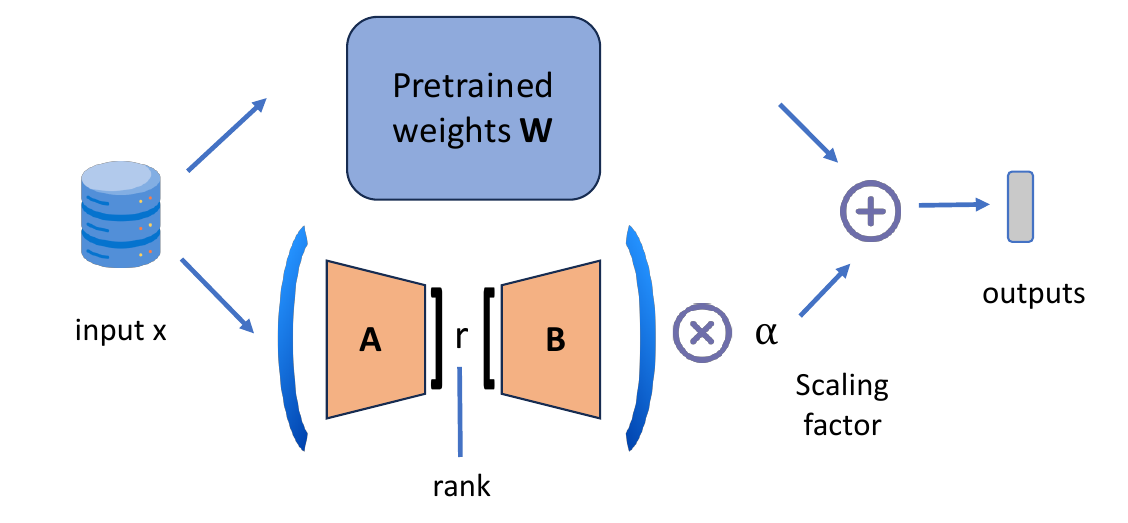}}

\caption{Schematic diagram of LoRA finetuning Mechanism. This figure demonstrates the LoRA fine-tuning process applied to a pre-trained model. Input $x$ is processed through low-rank matrices 
$\boldsymbol{A}$ and $\boldsymbol{B}$, which are the core components of the LoRA approach. These matrices modify the pre-trained weights $\boldsymbol{W}$ by a scaling factor $\alpha$, allowing for precise adjustments to the model without the need to retrain the entire network. This targeted fine-tuning strategy leads to efficient adaptation and output generation.
}

\vspace{-2.0em}

\label{fig:lora}
\end{center}
\end{figure}

\subsection{Low-Rank Adaptation for Vision-Language Models}

Low-Rank Adaptation (LoRA)~\cite{hu2022lora} presents a novel and efficient approach to fine-tuning deep learning models, particularly in the context of VLMs. This technique, by focusing on modifying only a small subset of model parameters, significantly reduces the computational overhead and memory requirements traditionally associated with training large neural networks. In the realm of vision-language tasks, where models must capture complex, multimodal relationships, LoRA offers a path to enhance model adaptability and performance without necessitating extensive computational resources.

LoRA operates by introducing trainable low-rank matrices into specific layers of a pre-trained model. These matrices serve as adapters, adjusting the model's behavior to better suit the task at hand with minimal interference to the pre-trained parameters. This approach is particularly beneficial in FL environments, where computational resources are dispersed and data privacy is of paramount importance. Figure~\ref{fig:lora} demonstrates the mechanism of LoRA. This methodology is premised on the addition of low-rank matrices, denoted as $\boldsymbol{A}$ and $\boldsymbol{B}$, which are applied to the input $x$ in conjunction with the original weight matrix $\boldsymbol{W}$. The key operation can be succinctly represented as

\begin{equation}
\label{eq:lora}
   \boldsymbol{W'} = (\boldsymbol{W} + \alpha . \boldsymbol{A}. \boldsymbol{B})\\
\end{equation}

where $\boldsymbol{W'}$ is the new updated weight matrix, while $\alpha$, the scaling factor and the rank $r$ are our hyperpamaters. The rank $r$ is a hyperparameter that controls the inner dimension of the matrices $\boldsymbol{A}$ and $\boldsymbol{B}$. Then the output can be obtained as:  $x.\boldsymbol{W'}$ during inference. This formulation effectively separates the low-rank matrices from the original weights, facilitating a unique adaptation mechanism that is both memory-efficient and flexible. Note that if $\boldsymbol{W} \in \mathbb{R}^{d \times k}$, then $\boldsymbol{A} \in \mathbb{R}^{d \times r}$ and $\boldsymbol{B} \in \mathbb{R}^{r \times k}$.

The adoption of LoRA within FL frameworks endows clients with the capacity for localized, efficient model updates, which is a cornerstone for collaborative learning that upholds the sanctity of data privacy. Recent studies have ventured into integrating LoRA specifically within the fully connected layers of deep neural networks in a FL context. For instance, Wu et al.~\cite{wu2024fedlora} examined the implementation of LoRA adapters to fine-tune fully connected layers, while Yi et al.~\cite{yi2024pfedlora} explored the prospects of personalized FL through parameter-efficient LoRA updates. These approaches underscore the growing interest and potential in enhancing model adaptability and efficiency, crucial for FL's scalability and practicality. Of late, parameter-efficient low-rank adapters have benefited from Neural Architecture Search~(NAS) techniques that can discover high-performing configurations in a search space of elastic LoRA adapters \cite{lonas2024, shears2024}.  

The application of LoRA to VLMs, such as the CLIP model, allows for nuanced and efficient adaptation to diverse datasets encountered in federated networks. By selectively updating parameters through low-rank matrices, the model can quickly adjust to new visual and textual information, enhancing its ability to generate accurate and contextually relevant outputs. This capability is especially valuable in scenarios where data distributions vary significantly across clients, a common challenge in FL settings.

Furthermore, LoRA's parameter-efficient fine-tuning aligns with the core principles of FL by enabling clients to contribute to a global model without the need for extensive computational power or bandwidth. This approach not only mitigates the risk of data privacy breaches but also democratizes access to advanced AI technologies, allowing entities with limited resources to participate in the collaborative development of robust, multimodal AI systems.


\begin{table}[ht]
\caption{The test accuracy (\%) of the image classification tasks in IID setting with $N=10$.}
\label{tab:iid_vertical}
\begin{center}
\begin{tabular}{c|cccc|c}
\toprule
\multicolumn{1}{c} {\textbf{Dataset}} & \textbf{FedFFT} & \textbf{FedLC} & \textbf{FedVM-LC} & \textbf{FedAA} & \textbf{Ours} \\
\midrule

\multirow{1}{*} {F-MNIST}       & 79.69 & 70.13 & 82.39 & 81.63 & \cellcolor{blue!10} \textbf{89.89} \\
                                
\multirow{1}{*} {CIFAR-10}      & 87.26 & 89.69 & 86.31 & 89.57 & \cellcolor{blue!10} \textbf{94.49} \\
                                
\multirow{1}{*} {CIFAR-100}     & 64.88 & 60.61 & 61.52 & 66.68 & \cellcolor{blue!10} \textbf{79.98} \\

\multirow{1}{*} {TINY}          & 59.77 & 60.30 & 59.77 & 64.46 & \cellcolor{blue!10} \textbf{75.99} \\
                                
\multirow{1}{*} {OxfordPets}    & 80.66 & 81.62 & 81.41 & 84.13 & \cellcolor{blue!10} \textbf{91.21} \\
                                
\multirow{1}{*} {Flowers102}    & 85.5 & 53.5 & 53.5 & 67.0 & \cellcolor{blue!10} \textbf{98.0} \\
                                
\multirow{1}{*} {Aircraft}      & 16.66 & 16.66 & 19.29 & 21.49 & \cellcolor{blue!10} \textbf{56.14} \\

\multirow{1}{*} {Cars}          & 44.85 & 46.81 & 44.85 & 50.0 & \cellcolor{blue!10} \textbf{85.53} \\

\multirow{1}{*} {DTD}           & 55.30 & 44.69 & 43.18 & 46.21 & \cellcolor{blue!10} \textbf{85.60} \\

\multirow{1}{*} {EuroSAT}       & 86.11 & 50.15 & 89.04 & 78.08 & \cellcolor{blue!10} \textbf{95.67} \\

\multirow{1}{*} {FER2013}       & 41.14 & 37.40 & 52.61 & 66.33 & \cellcolor{blue!10} \textbf{74.06} \\

\multirow{1}{*} {Caltech101}    & 84.65 & 79.36 & 78.83 & 83.06 & \cellcolor{blue!10} \textbf{97.88} \\
                                
\multirow{1}{*} {Food101}       & 77.30 & 77.30 & 77.30 & 76.99 & \cellcolor{blue!10} \textbf{82.62} \\
                                
\multirow{1}{*} {Country211}    & 18.56 & 18.69 & 18.56 & 18.06 & \cellcolor{blue!10} \textbf{27.94} \\
                                
\multirow{1}{*} {SUN397}        & 65.97 & 66.58 & 65.64 & 68.67 & \cellcolor{blue!10} \textbf{81.22} \\

\multirow{1}{*} {R-SST2}          & \cellcolor{blue!10} \textbf{97.42} & 75.53 & 80.68 & 91.41 & 82.40 \\

\bottomrule
\end{tabular}

\vspace{-2.0em}

\end{center}
\end{table}

\begin{table}[ht]
\caption{The test accuracy (\%) of the image classification tasks in practical non-IID setting with $N = 10$ (method names omitted the word \textbf{"Fed" }for saving space).}
\label{tab:prac_vertical}
\begin{center}
\begin{tabular}{c|c|cccc|c}
\toprule
\multicolumn{1}{c} {\textbf{Dataset}} & \multicolumn{1}{|c|} {\boldmath{$\beta$}} & \textbf{FFT} & \textbf{LC} & \textbf{VM-LC} & \textbf{AA} & \textbf{Ours} \\
\midrule

\multirow{2}{*} {F-MNIST}  
                                & $0.1$ & 63.88 & 63.88 & 63.88 & 59.96 & \cellcolor{blue!10} \textbf{84.30} \\
                                & $1.0$ & 75.36 & 63.88 & 72.12 & 77.82 & \cellcolor{blue!10} \textbf{89.38} \\ 
                                
\midrule                                
\multirow{2}{*} {CIFAR-10} 
                                & $0.1$ & 79.73 & 85.42 & 79.73 & 81.10 & \cellcolor{blue!10} \textbf{93.06} \\
                                & $1.0$ & 88.27 & 88.27 & 88.27 & 91.06 & \cellcolor{blue!10} \textbf{94.39} \\ 
                                
\midrule                                
\multirow{2}{*} {CIFAR-100}      
                                & $0.1$ & 60.61 & 60.74 & 60.61 & 62.10 & \cellcolor{blue!10} \textbf{74.04} \\
                                & $1.0$ & 62.31 & 62.31 & 62.31 & 66.45 & \cellcolor{blue!10} \textbf{77.09} \\ 
                                
\midrule
\multirow{2}{*} {TINY}      
                                & $0.1$ & 59.77 & 59.77 & 59.77 & 61.52 & \cellcolor{blue!10} \textbf{71.67} \\
                                & $1.0$ & 59.59 & 59.59 & 59.59 & 61.90 & \cellcolor{blue!10} \textbf{71.29} \\ 
                                
\midrule                                
\multirow{2}{*} {OxfordPets}
                                & $0.1$ & 80.66 & 81.21 & 80.66 & 77.46 & \cellcolor{blue!10} \textbf{87.13} \\
                                & $1.0$ & 80.66 & 81.00 & 80.66 & 81.82 & \cellcolor{blue!10} \textbf{90.74} \\ 
                                
\midrule                                
\multirow{2}{*} {Flowers102}  
                                & $0.1$ & 53.5 & 54.0 & 53.5 & 47.5 & \cellcolor{blue!10} \textbf{91.99} \\ 
                                & $1.0$ & 65.11 & 66.46 & 65.11 & 67.86 & \cellcolor{blue!10} \textbf{95.11} \\ 
                                
\midrule                                
\multirow{2}{*} {Aircraft}      
                                & $0.1$ & 16.66 & 16.66 & 16.66 & 18.42 & \cellcolor{blue!10} \textbf{44.29} \\ 
                                & $1.0$ & 17.84 & 17.84 & 19.25 & 21.05 & \cellcolor{blue!10} \textbf{44.75} \\ 
                                
\midrule                                
\multirow{2}{*} {Cars}  
                                & $0.1$ & 44.85 & 46.81 & 44.85 & 50.0 & \cellcolor{blue!10} \textbf{86.76} \\
                                & $1.0$ & 47.51 & 47.85 & 47.51 & 52.48 & \cellcolor{blue!10} \textbf{75.84} \\

\midrule                                
\multirow{2}{*} {DTD}          
                                & $0.1$ & 40.15 & 40.90 & 40.15 & 40.90 & \cellcolor{blue!10} \textbf{65.15} \\
                                & $1.0$ & 48.84 & 41.13 & 43.26 & 43.61 & \cellcolor{blue!10} \textbf{70.92} \\

\midrule                                
\multirow{2}{*} {EuroSAT}
                                & $0.1$ & 43.20 & 49.38 & 43.20 & 37.96 & \cellcolor{blue!10} \textbf{92.43} \\
                                & $1.0$ & 65.38 & 45.18 & 68.59 &68.96 & \cellcolor{blue!10} \textbf{94.44} \\

\midrule                                
\multirow{2}{*} {FER2013}   
                                & $0.1$ & 39.40 & 37.15 & 37.15 & 40.39 & \cellcolor{blue!10} \textbf{62.84} \\
                                & $1.0$ & 40.41 & 44.88 & 40.67 & 44.78 & \cellcolor{blue!10} \textbf{64.79} \\          
\midrule                                
\multirow{2}{*} {Caltech101}          
                                & $0.1$ & 78.30 & 78.30 & 78.30 & 79.89 & \cellcolor{blue!10} \textbf{94.58} \\
                                & $1.0$ & 88.18 & 88.24 & 88.18 & 87.49 & \cellcolor{blue!10} \textbf{95.96} \\ 

\midrule                                
\multirow{2}{*} {Food101}       
                                & $0.1$ & 77.30 & 77.30 & 77.30 & 74.79 & \cellcolor{blue!10} \textbf{78.02} \\
                                & $1.0$ & 79.59 & 79.59 & 79.59 & 79.62 & \cellcolor{blue!10} \textbf{82.79} \\
                                
\midrule                                
\multirow{2}{*} {Country211}    
                                
                                & $0.1$ & 18.56 & 18.56 & 18.56 & 17.68 & \cellcolor{blue!10} \textbf{25.73} \\ 
                                & $1.0$ & 15.13 & 15.13 & 15.13 & 15.20 & \cellcolor{blue!10} \textbf{23.05} \\ 
                                
\midrule                                
\multirow{2}{*} {SUN397}        
                                & $0.1$ & 65.64 & 65.64 & 65.64 & 66.78 & \cellcolor{blue!10} \textbf{77.58} \\
                                & $1.0$ & 59.54 & 61.56 & 59.54 & 61.12 & \cellcolor{blue!10} \textbf{73.39} \\ 
                                
\midrule                                
\multirow{2}{*} {R-SST2}          
                                & $0.1$ & 97.42 & 97.42 & \cellcolor{blue!10} \textbf{97.85} & 92.27 & 88.84 \\
                                & $1.0$ & 60.19 & 60.19 & 60.19 & 54.31 & \cellcolor{blue!10} \textbf{68.00} \\ 
                                
\bottomrule
\end{tabular}
\end{center}

\vspace{-2.0em}

\end{table}

\section{Methodology}

In this section, we present the methodology employed in our study on federated fine-tuning learning for VLMs. We outline the framework and techniques used to enable collaborative training across distributed devices while incorporating fine-tuning for enhanced performance and privacy preservation.

    
\subsection{Baseline fine-tuning methods}

\subsubsection{Full fine-tuning (FFT)} 
In full fine-tuning, we update the whole CLIP model, i.e., both image and text encoder, by minimizing the cross-entropy loss on labeled downstream data.

\subsubsection{Linear classifier (LC)} 
We learn a linear classifier $h_{\text{class}} \in \mathbb{R}^{d \times k}$ on top of frozen image embedding $\bar{f}(I)$ by minimizing the cross-entropy on downstream data~\cite{radford2021clip}. The text encoder is not involved during this process. As Wortsman et al.~\cite{wortsman2022wiseft} have suggested, instead of randomly initialing the linear classifier, the zero-shot weights of the CLIP's text encoder are applied to it.

\subsubsection{Fine-tune vision model with linear classifier (VM-LC)} 
Inspired by Goyal et al.~\cite{goyal2022flyp}, in VM-LC, we update both a linear head and the parameters of the image encoder $\Theta_{\text{img}}$ (initialized at the pre-trained value) by minimizing the cross-entropy loss on labeled downstream data. The text encoder is not involved during this process. Similar to the previous method, we initialize the linear head the same as the pre-trained weights of the text encoder.

\subsubsection{Attention Adapter (AA)\texorpdfstring{~\cite{lu2023fedclip}}{}}

Here, we integrate an attention adapter after the image encoder to perform a fine-tuning process. The whole CLIP model is kept frozen during training.

\subsection{Approach}

Our approach innovatively adapts the CLIP model, leveraging its powerful pre-trained representation for image and text embeddings within a FL framework. At the core of our strategy lies the integration of LoRA fine-tuning, which allows for efficient model adaptation across distributed clients while preserving the intrinsic knowledge captured during CLIP's extensive pre-training.

\begin{figure}[ht]
\begin{center}
    \begin{subfigure}[b]{0.325\linewidth}
        \includegraphics[width=\linewidth]{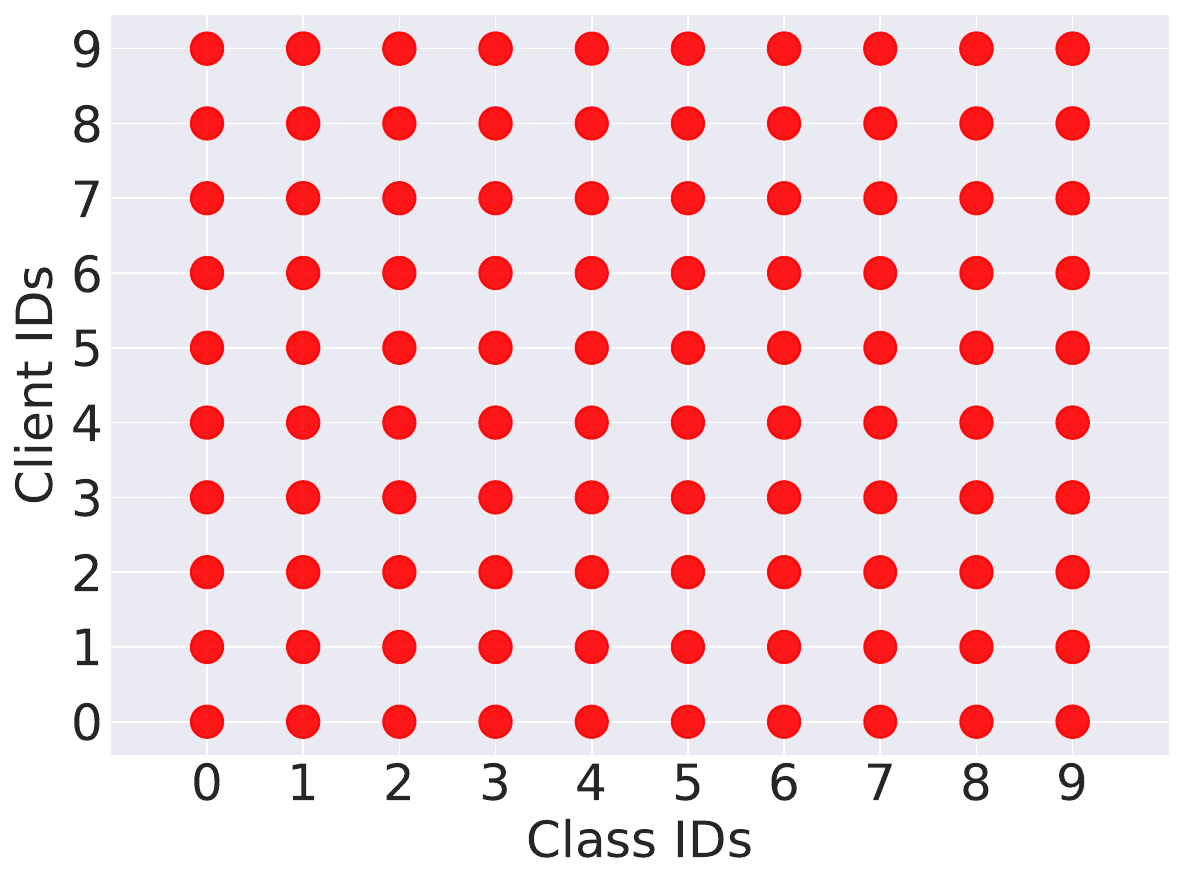}
        \caption{IID}
        \label{fig:cifar10_iid}
    \end{subfigure} \hspace{0em}%
    \hfill
    \begin{subfigure}[b]{0.325\linewidth}
        \includegraphics[width=\linewidth]{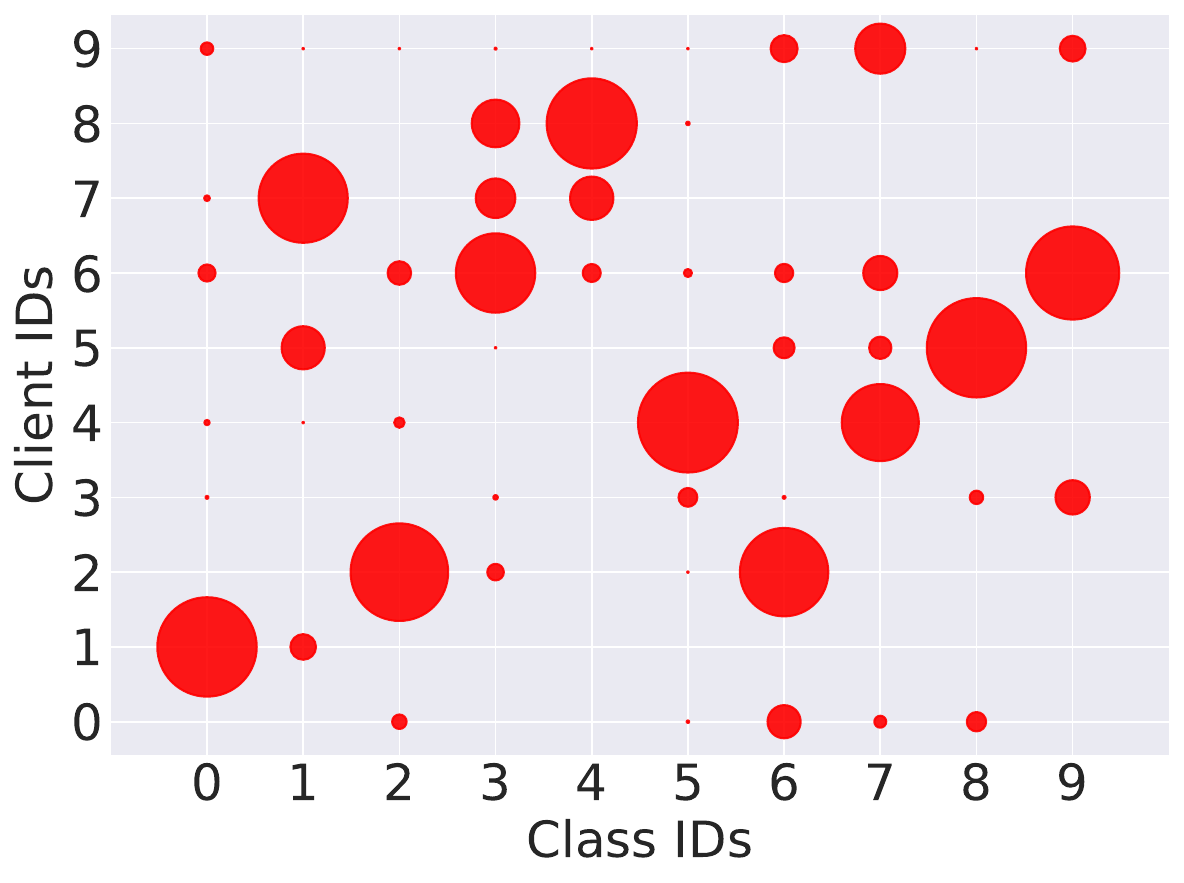}
        \caption{$\beta = 0.1$}
        \label{fig:cifar10_dir05}
    \end{subfigure} \hspace{0em}%
    \hfill
    \begin{subfigure}[b]{0.325\linewidth}
        \includegraphics[width=\linewidth]{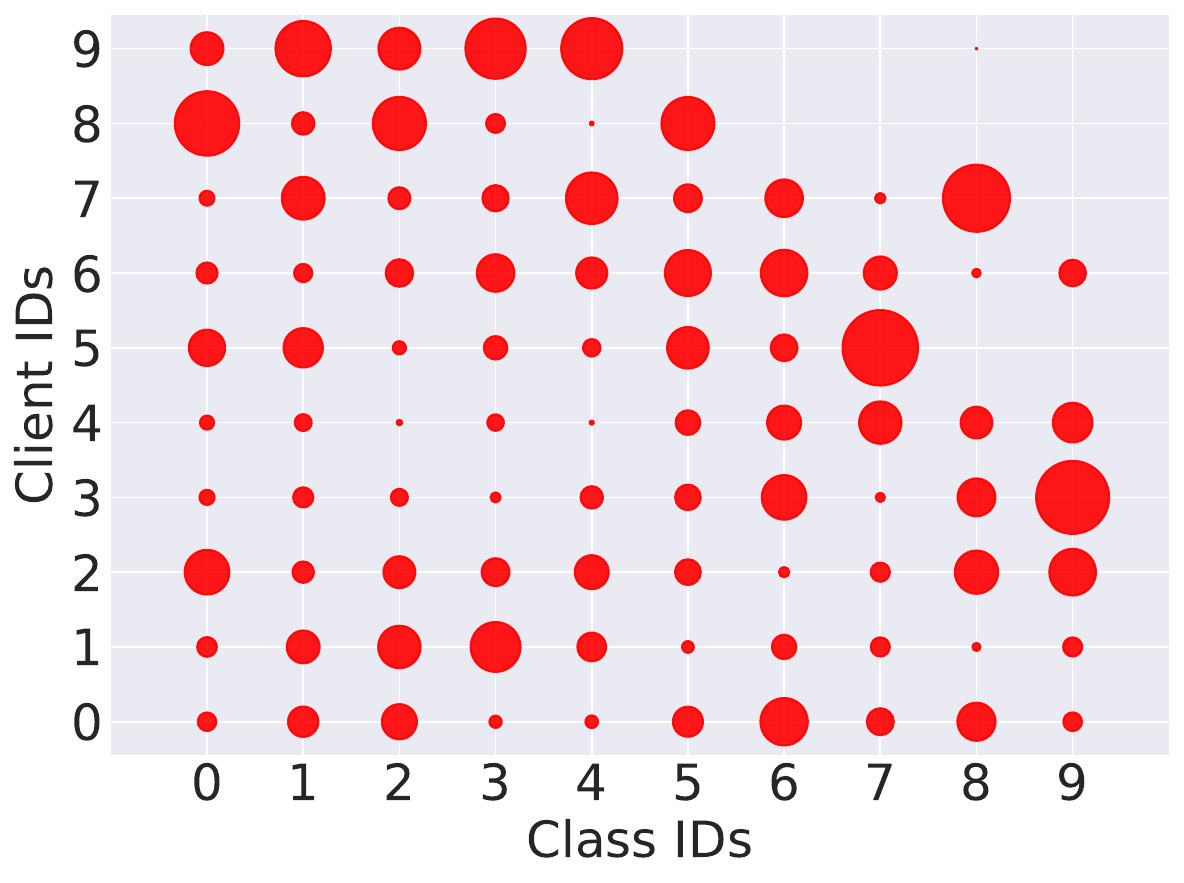}
        \caption{$\beta = 1.0$}
        \label{fig:cifar10_dir10}
    \end{subfigure}
    \caption{The data distribution of all clients on CIFAR-10 in IID and non-IID setting with varying $\beta$. The size of a circle means the number of data samples.}
    \label{fig:cifar10_distribution}
\end{center}

\vspace{-2.0em}

\end{figure}

Given a pre-trained CLIP backbone, the input to the text encoder is crafted in the form "a photo of a [class]", where [class] denotes the category of the image within the dataset~\cite{radford2021clip}. Leveraging CLIP's ability to predict image-text matchings, the classification loss and logits can be calculated by aligning the embedding spaces for images and textual descriptions corresponding to "[class]".

Formally, let \(\mathcal{I}(\cdot)\) denote the image encoder, and \(\mathcal{T}(\cdot)\) the text encoder, which remains fixed except for the low-rank adaptations through LoRA. For a given class \(i\), let \(\mathbf{t}_i\) be the word embedding vector of the class label \( [\text{class}]_i\), with \(i \in [1, K]\) and \(K\) representing the number of classes.

We utilize the cosine similarity function \(\cos[\cdot|\cdot]\), as employed by CLIP, to measure the similarity between the image and text features. The classification prediction probabilities and logits for an image \(\mathbf{x}\) against a class \(i\) are calculated as follows:

\begin{equation}
\label{eq:cosine_similarity}
    p(\textbf{y} = i | \textbf{x}) = \frac{\exp(\cos[\mathcal{I}(\textbf{x}), \mathcal{T}(\mathbf{t}_i)) / \tau])}{\sum_{j=1}^K \exp(\cos[\mathcal{I}(\textbf{x}), \mathcal{T}(\mathbf{t}_j)) / \tau])}
\end{equation}

where \(\tau\) is the temperature coefficient learned by CLIP.
The text encoder's LoRA parameters are the only variables updated during local backpropagation and subsequently aggregated by the federated server. The image encoder, the main body of the text encoder, and the word embedding \(\mathbf{t}_i\) is kept frozen to preserve the foundational knowledge embedded within the pre-trained CLIP model.

In this framework, the loss function for a batch of images and their associated classes is the cross-entropy loss between the predicted probabilities \(p(\textbf{y} = i | \textbf{x})\) and the ground truth labels. By minimizing this loss, we perform selective fine-tuning of the model's textual understanding to better align with the domain-specific data encountered in each client's local dataset while capitalizing on the pre-trained multimodal embeddings of the CLIP model.


\begin{figure}[ht]


\begin{center}
    \begin{subfigure}[b]{0.493\linewidth}
        \includegraphics[width=\linewidth]{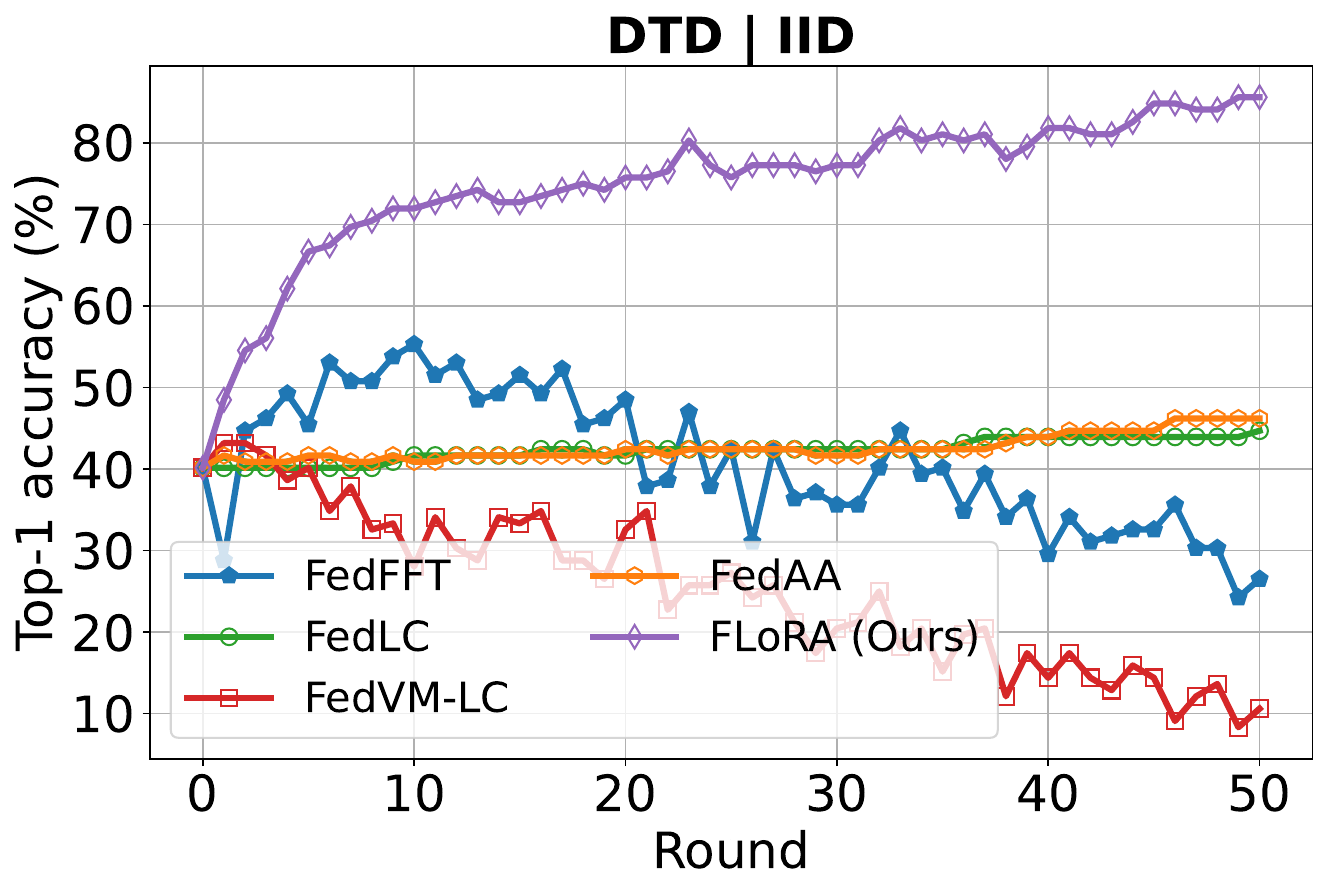}
        \label{fig:d47_iid}
    \end{subfigure} \hspace{0em}%
    \hfill
    \begin{subfigure}[b]{0.493\linewidth}
        \includegraphics[width=\linewidth]{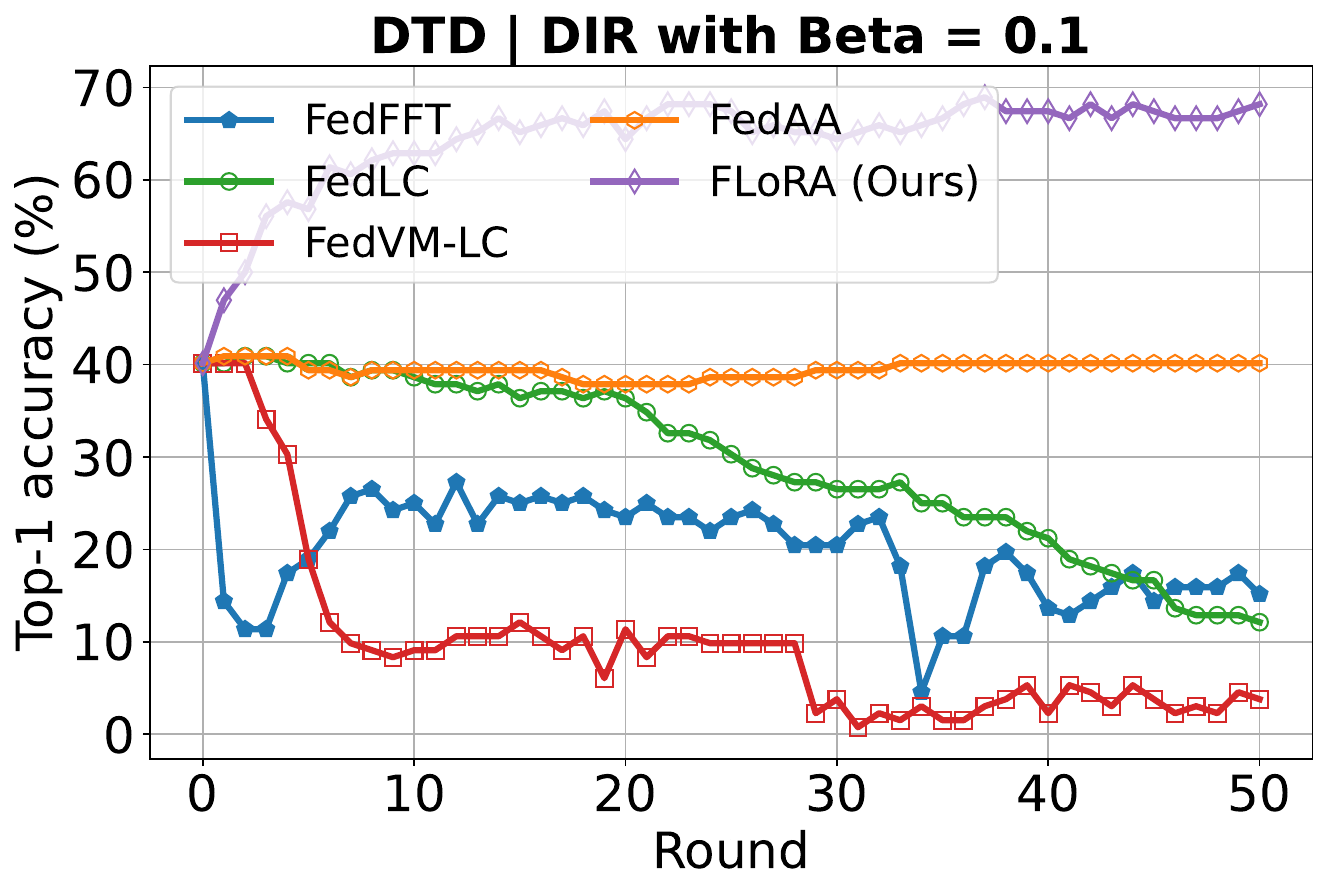}
        \label{fig:d47_dir}
    \end{subfigure} \hspace{0em}%
    
    \caption{Learning curve for DTD dataset in different settings.}
    \label{fig:d47_learning_curve}


\end{center}

\vspace{-2.0em} 

\end{figure}

\begin{figure}[ht]


\begin{center}
    

    \begin{subfigure}[b]{\linewidth}
        \includegraphics[width=\linewidth]{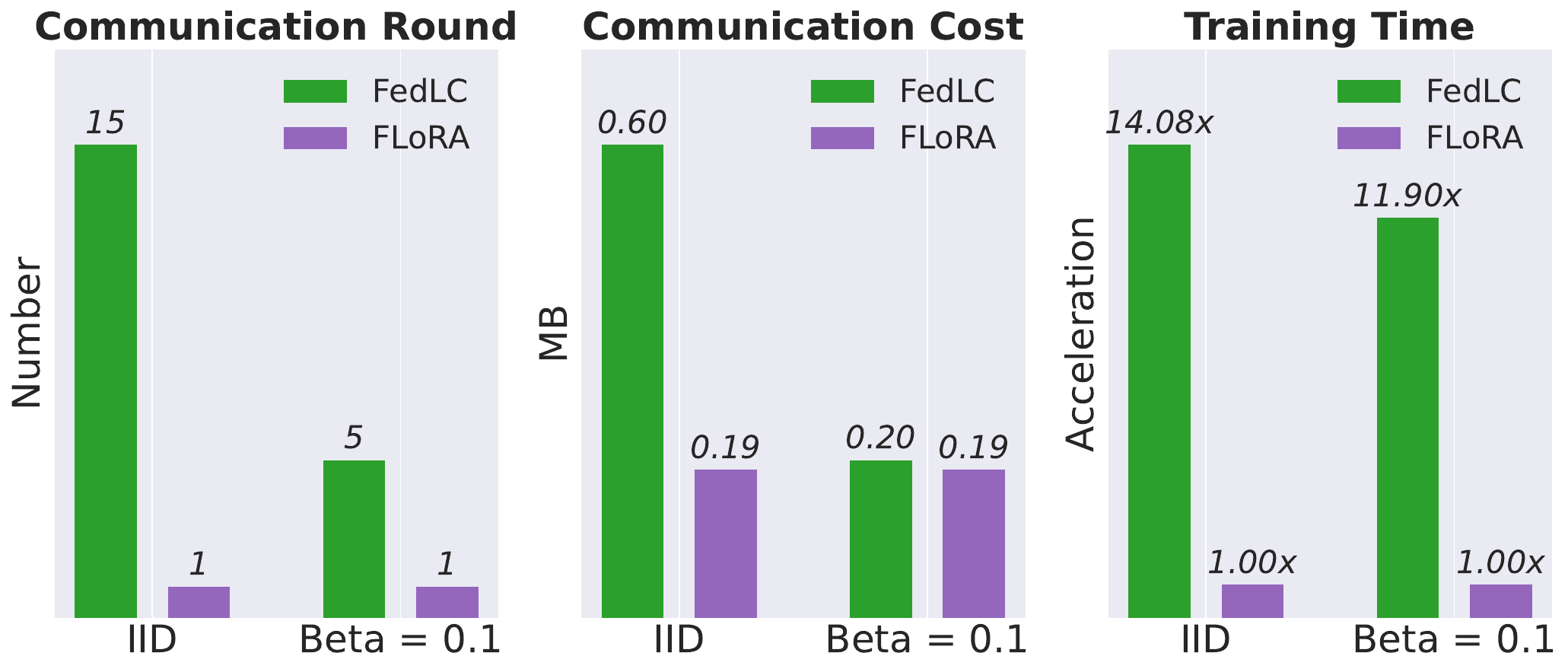}
        \caption{Comparison of the best two approaches in CIFAR-10 that reach $89 \%$ and $85 \%$ target accuracy for IID and Non-IID respectively.}
        \label{fig:cifar10_overhead_iid_dir}
    \end{subfigure} \hspace{0em}%
    
    \vspace{0.0em} 
    
    \begin{subfigure}[b]{\linewidth}
        \includegraphics[width=\linewidth]{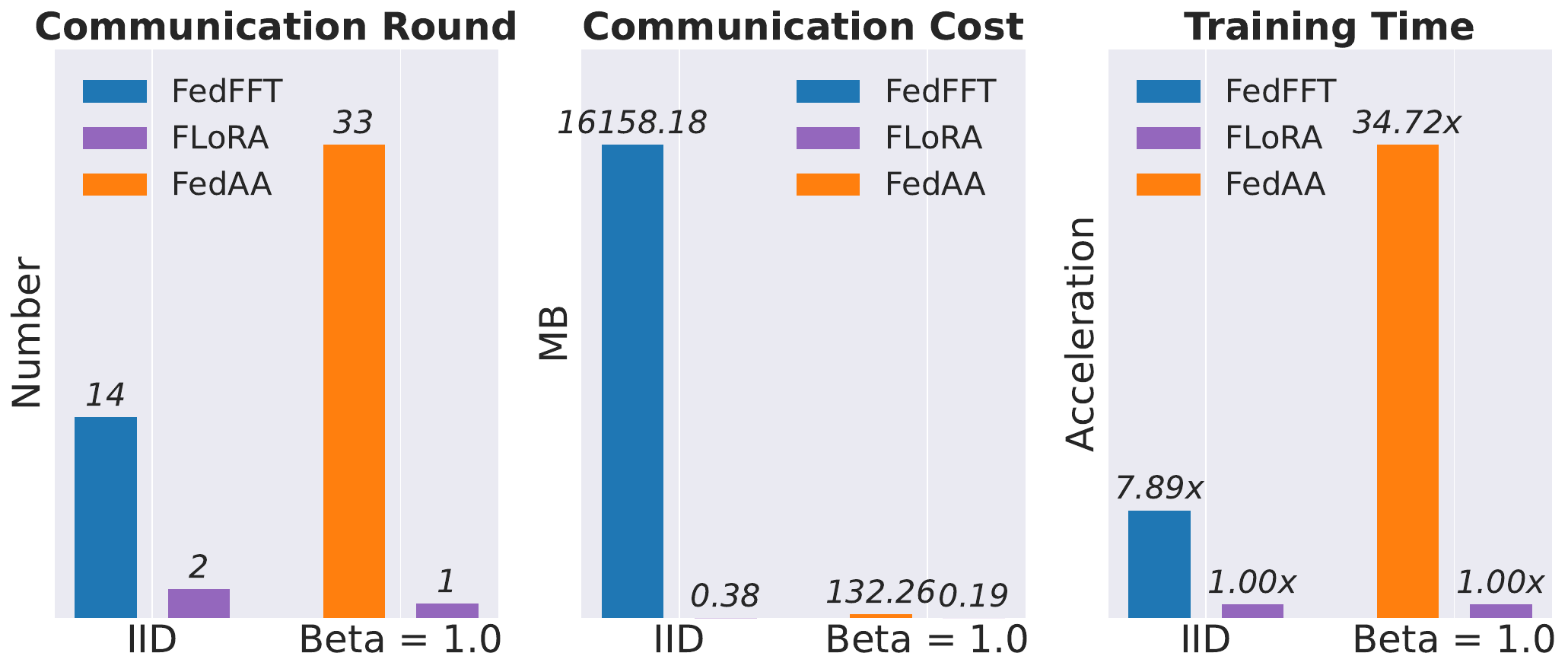}
        \caption{Comparison of the best two approaches in Flowers102 that reach $85 \%$ and $67 \%$ target accuracy for IID (FedFFT and \proj) and Non-IID (FedAA and \proj) respectively.}
        \label{fig:flowers_overhead_iid_dir}
    \end{subfigure} \hspace{0em}%

    

    \vspace{0.0em} 
    
    \begin{subfigure}[b]{\linewidth}
        \includegraphics[width=\linewidth]{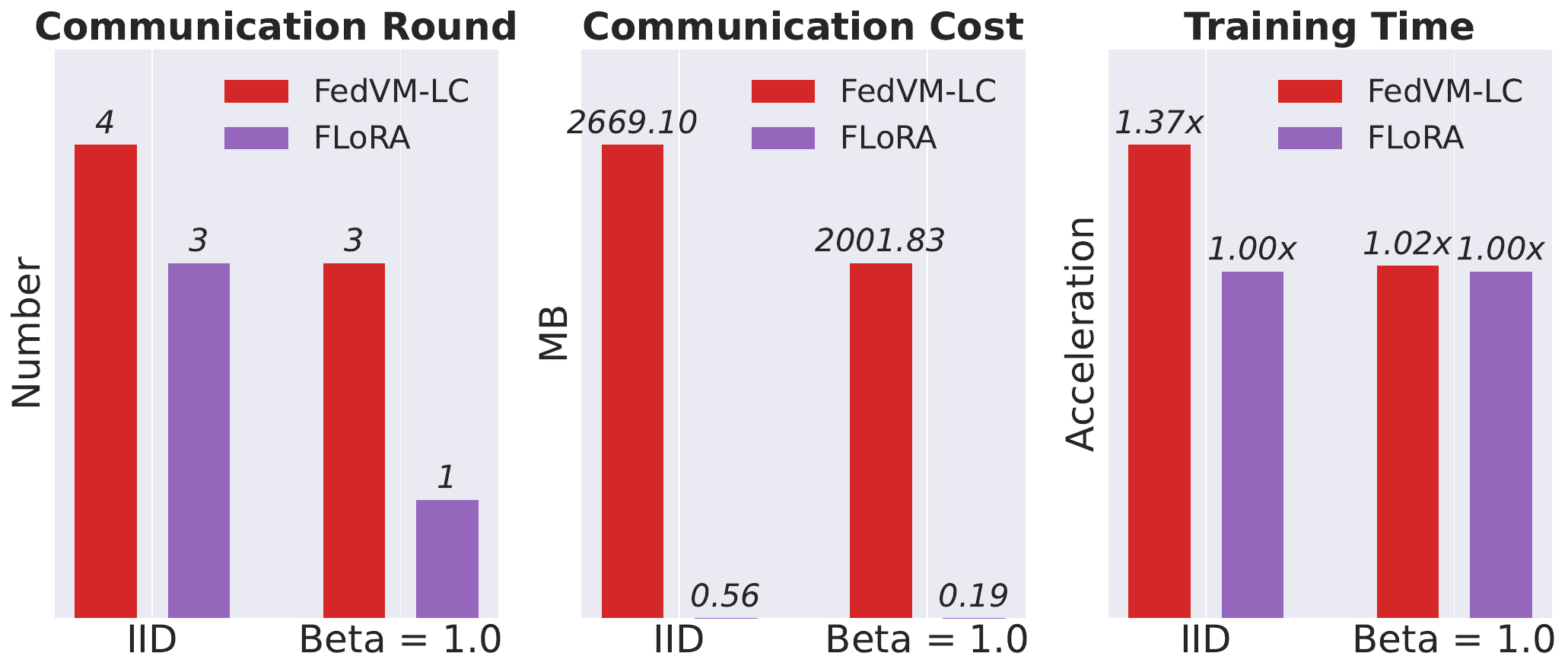}
        \caption{Comparison of the best two approaches in EuroSAT that reach $89 \%$ and $68 \%$ target accuracy for IID and Non-IID respectively.}
        \label{fig:eurosat_overhead_iid_dir}
    \end{subfigure} \hspace{0em}%

    \vspace{0.0em}
    
    \caption{Comparison of computation and communication overhead in selected datasets to reach target accuracies. \proj requires fewer communication rounds while having significantly less communication cost and training time compared to the second best method.}
    \label{fig:overhead}
\end{center}

\vspace{-2.0em}

\end{figure}


\begin{figure}[ht]

\vspace{-2.0em}

\begin{center}
    \centerline{\includegraphics[width=0.3\textwidth]{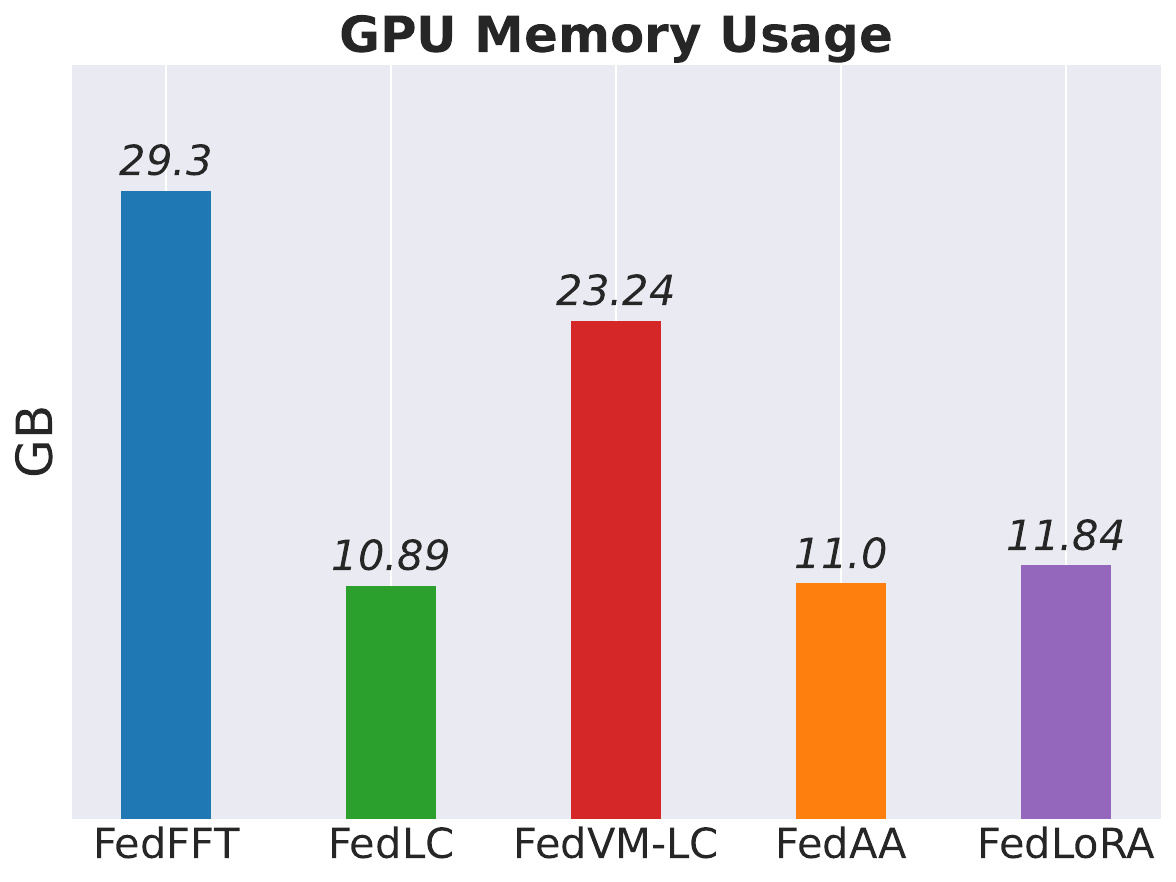}}
    
    \caption{Memory usage of all methods.}
    \label{fig:memory_all}
\end{center}

\vspace{-2.0em}

\end{figure}

\section{Experiments}

\begin{table}[ht]
\caption{The number of parameters and size of transferred models for each method. For \textbf{FedLC} and \textbf{FedVM-LC}, there are ranges of trainable parameters and size depending on the dataset.}
\label{tab:communication}
\begin{center}
\begin{tabular}{c|cc}
\toprule
& \multicolumn{1}{c}{Number of trainable parameters} & \multicolumn{1}{c}{Size (MB)} \\ 
\midrule
\textbf{FedFFT}     & 151,277,313 & 577.078 \\

\textbf{FedLC}      & 1026 -- 203661 & 0.004 -- 0.777  \\

\textbf{FedVM-LC}   & 87,457,026 -- 87,659,661 & 333.622 -- 334.395 \\

\textbf{FedAA}      & 525,312 & 2.004 \\

\midrule

\textbf{Ours}       & 24,576 & 0.094 \\
\bottomrule
\end{tabular}
\end{center}


\end{table}

We conduct comprehensive experiments to evaluate the effectiveness of our federated fine-tuning approach.

\subsection{Datasets and Experimental Setting}

\subsubsection{Datasets}
We use publicly available vision datasets: Fashion-MNIST~(F-MNIST)~\cite{xiao2017fmnist}, CIFAR-10, CIFAR-100~\cite{krizhevsky2009cifar}, Tiny-ImageNet~(TINY)~\cite{chrabaszcz2017tiny}, Oxford-IIIT Pet~(OxfordPets)~\cite{parkhi2012pets}, Oxford 102 Flower~(Flowers102)~\cite{nilsback2008flowers}, FGVC-Aircraft~(Aircraft)~\cite{maji2013aircraft}, Stanford Cars~(Cars)~\cite{krause2013cars}, Describable Textures Dataset~(DTD)~\cite{cimpoi14dtd}, EuroSAT~\cite{helber2019eurosat,helber2018introducing}, FER2013~\cite{dumitru13fer}, Caltech101~\cite{feifei2004caltech101}, Food101~\cite{bossard2014food}, Country211~\cite{radford2021clip}, SUN397~\cite{xiao2010sun1,xiao2014sun2}, and Rendered SST2~(R-SST2)~\cite{radford2021clip}.

To simulate the practical pFL setting, we evaluate the model on the same testing data, $20\%$ of the total dataset. The remaining $80\%$ data is used for training which is partitioned for all clients in different settings.

\subsection{Experimental setting}

We train all methods for $50$ rounds with one local epoch. Total number of client is $N=10$, the sample rate $\rho = 1$ with batch size equal to $128$. For Adam optimizer, learning rate is $5 \times 10^{-5}$,  $\epsilon = 10^{-6}$, and $0.2$ weight decay. All the experiments are done using $1$ GPU A6000 with $48$ GB. The pre-trained CLIP model is utilized with Vit-B/32~\cite{dosovitskiy2021vit} as the base image encoder. For LoRA paramters, we set the rank $r = 2 $ and scaling factor $\alpha=32$.
We create two popular statistically heterogeneous settings to simulate the FL environment: balanced IID~\cite{mcmahan2017fedavg} and label skew practical non-IID~\cite{lin2020feddf,li2021moon,li2022niidbench}.


\subsubsection{Homogeneous setting}
Data on each client are separated in IID fashion~\cite{mcmahan2017fedavg}. 

\subsubsection{Practical heterogeneous setting}
The second scenario is the practical heterogeneous setting~\cite{lin2020feddf,li2021moon}, which is controlled by the Dirichlet distribution, denoted as $Dir(\beta)$. The smaller the $\beta$ is, the more heterogeneous the setting is. We set $\beta = 0.1$ and $\beta = 1.0$ for the heterogeneous setting~\cite{lin2020feddf,wang2020fednova}. Figure~\ref{fig:cifar10_distribution} illustrates the distribution of training data on CIFAR-10 dataset.

\subsubsection{Evaluation Metrics}
Our evaluation focuses on task-specific performance metrics such as top-1 accuracy. Additionally, we analyze communication efficiency, memory consumption and few-shot evaluation.

\subsection{Learning Analysis}

In our comprehensive assessment of federated learning methods for image classification, \proj—our proposed method—stands out for its efficient parameter updates and effective learning from decentralized data. Empirical evaluations across diverse datasets have demonstrated \proj's superior performance, particularly in IID settings with $N=10$ clients, and its robust adaptability to varied classification tasks. Notably, except for the binary classification dataset R-SST2, \proj outshines other baseline methods, as depicted in Table~\ref{tab:iid_vertical}.


In Table \ref{tab:prac_vertical}, we delve into the performance of various FL methods under non-IID data distribution, which presents a more practical and challenging scenario. The table reflects test accuracy percentages for different values of $\beta$, where $\beta$ represents the degree of data distribution skewness among the clients. We aim to evaluate the resilience of each method to data heterogeneity, an aspect critical to FL applications.

The results across datasets highlight the varying ability of each FL method to cope with non-IID data. Our method demonstrates consistent performance superiority across almost all datasets and $\beta$ settings, emphasizing its robustness and suitability for practical FL deployments.

The learning curves for the DTD dataset, as visualized in Figure~\ref{fig:d47_learning_curve}, corroborate these findings, where \proj achieves faster convergence and maintains high accuracy with significantly fewer communication rounds compared to other methods. This rapid and robust performance, particularly in non-IID conditions, highlights \proj's potential for practical federated deployments, offering a well-suited solution for diverse and distributed data environments. Notice that all of the methods start at almost the same accuracy (as in zero-shot classification thanks to the strong generalization of the pre-trained CLIP model), but other time only \proj show significant improvement and maintain it's performance through the training.

\begin{table}[ht]
\caption{The distribution of classes across clients in few short learning pathalogical experiment.}
\label{tab:fs_classes_vertical}
\begin{center}
\begin{tabular}{c|cc|cc}
\toprule
& & & \multicolumn{2}{c} {\textbf{\# classes of each client}} \\
\multicolumn{1}{c|} {\textbf{Dataset}} & \multicolumn{1}{c} \boldmath{$\Sigma$} \textbf{class} & {\boldmath{$N$}}  & \boldmath{$1^{th} \to {(N-1)}^{th}$} & \boldmath{$N^{th}$} \\
\midrule

\multirow{1}{*} {F-MNIST}       & 10 & 5 & 2 & 2 \\ 

\multirow{1}{*} {CIFAR-10}      & 10 & 5 & 2 & 2 \\ 
                                
\multirow{1}{*} {CIFAR-100}     & 100 & 10 & 10 & 10 \\ 

\multirow{1}{*} {TINY}          & 200 & 10 & 20 & 20 \\   

\multirow{1}{*} {OxfordPets}    & 37 & 6 & 6 & 7 \\       

\multirow{1}{*} {Flowers102}    & 102 & 6 & 17 & 17 \\   

\multirow{1}{*} {Aircraft}      & 100 & 10 & 10 & 10 \\   

\multirow{1}{*} {Cars}          & 196 & 7 & 28 & 28 \\ 

\multirow{1}{*} {DTD}           & 47 & 7 & 6 & 11 \\ 

\multirow{1}{*} {EuroSAT}       & 10 & 5 & 2 & 2 \\ 

\multirow{1}{*} {FER2013}       & 7 & 3 & 2 & 3 \\ 

\multirow{1}{*} {Caltech101}    & 101 & 10 & 10 & 11 \\ 

\multirow{1}{*} {Food101}       & 101 & 10 & 10 & 11 \\     

\multirow{1}{*} {Country211}    & 211 & 10 & 21 & 22 \\   

\multirow{1}{*} {SUN397}        & 397 & 10 & 39 & 46 \\   

\multirow{1}{*} {SST2}          & 2 & 2 & 1 & 1 \\


\bottomrule
\end{tabular}
\end{center}


\end{table}

\begin{figure}[ht]
\begin{center}

    \begin{subfigure}[b]{0.493\linewidth}
        \includegraphics[width=\linewidth]{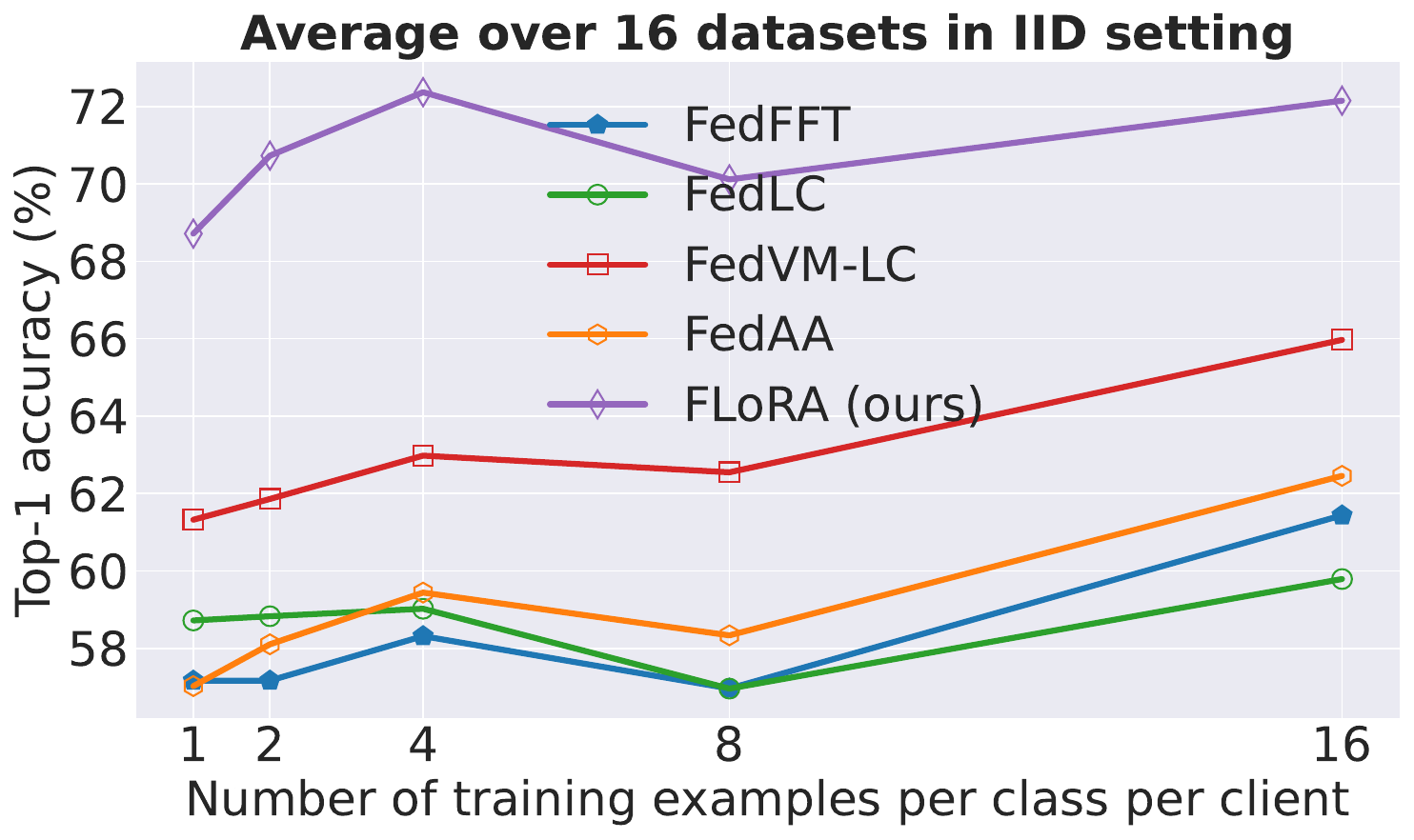}
        \caption{IID}
        \label{fig:average_iid_fs}
    \end{subfigure} \hspace{0em}%
    \hfill
    \begin{subfigure}[b]{0.493\linewidth}
        \includegraphics[width=\linewidth]{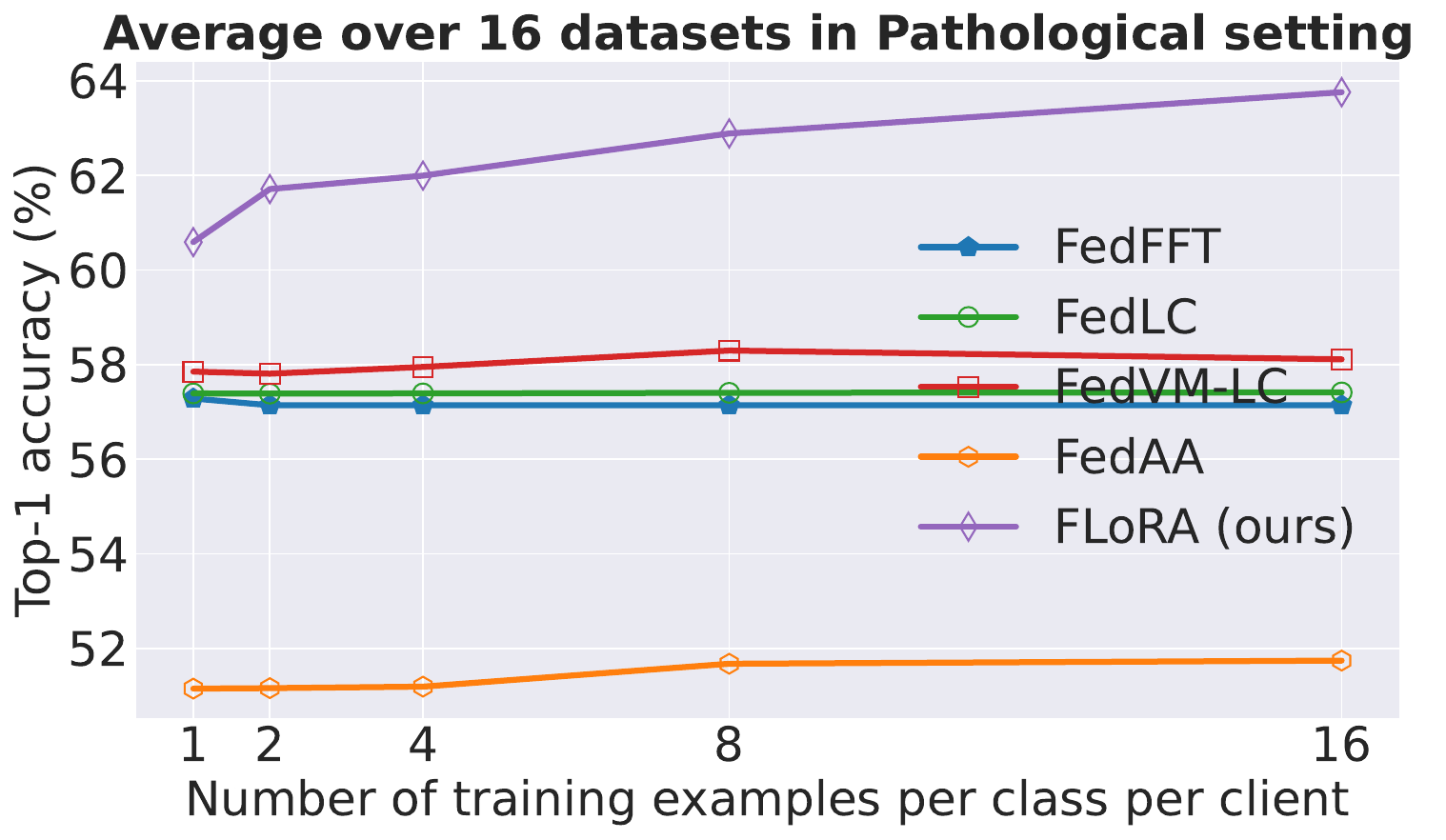}
        \caption{Pathological non-IID}
        \label{fig:average_pat_fs}
    \end{subfigure} \hspace{0em}%


    \caption{Average results of few-shot learning on various datasets.}
    \label{fig:fs_avg}
    
    \vspace{-2.0em}
    
\end{center}
\end{figure}

\begin{figure}[ht]

\begin{center}
    \begin{subfigure}[b]{0.492\linewidth}
        \includegraphics[width=\linewidth]{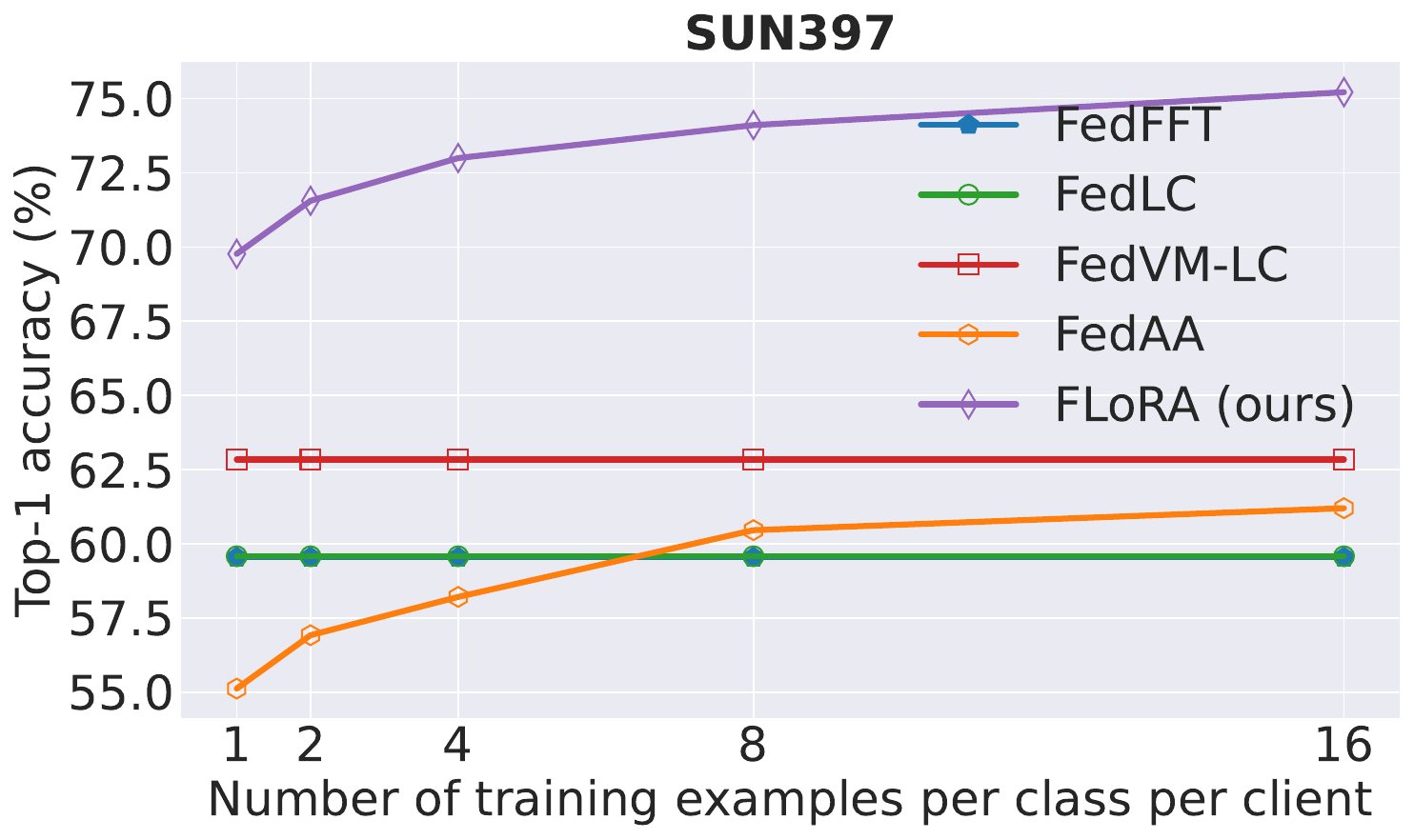}
        \caption{IID}
        \label{fig:sun397_iid_fs}
    \end{subfigure} \hspace{0em}%
    \hfill
    \begin{subfigure}[b]{0.492\linewidth}
        \includegraphics[width=\linewidth]{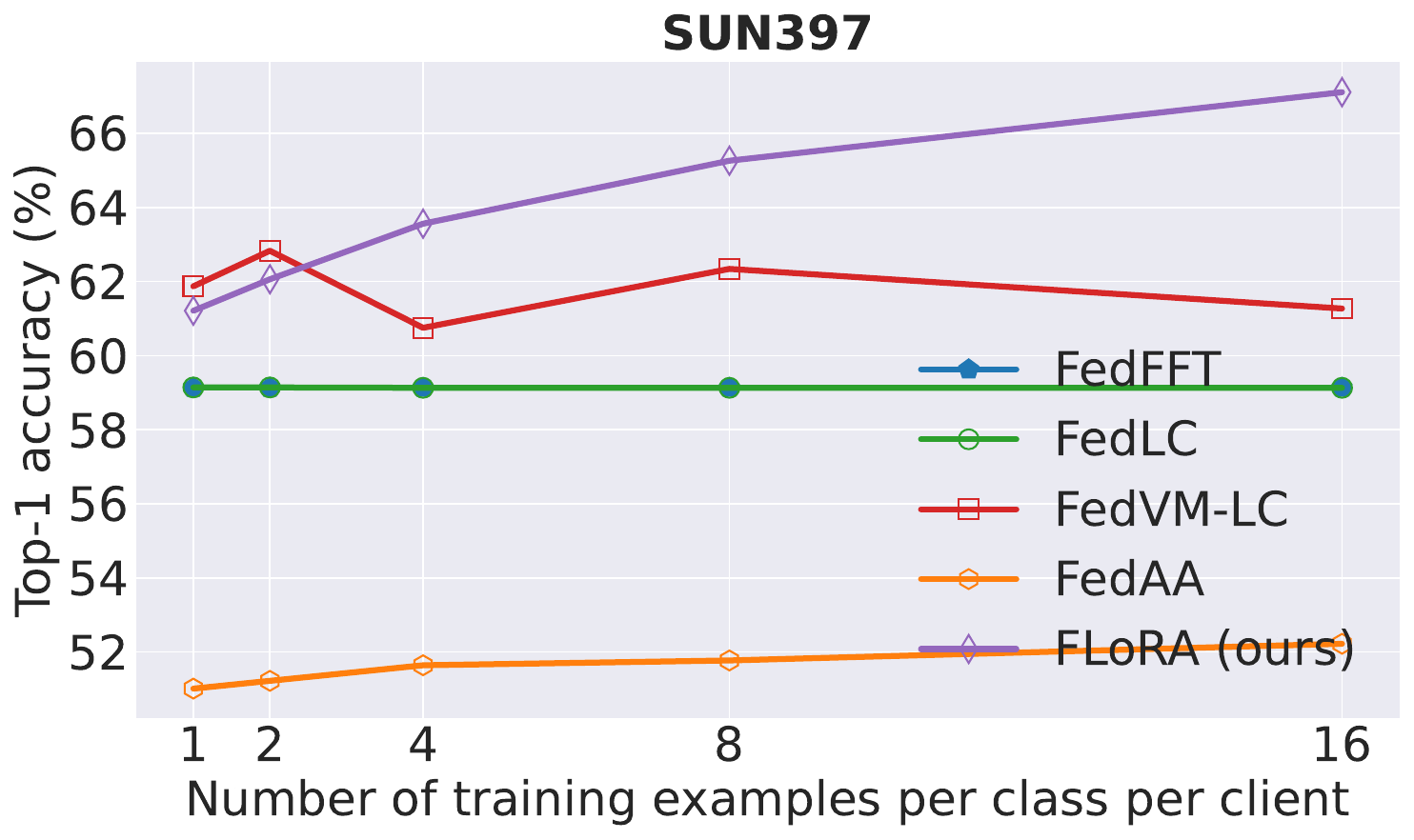}
        \caption{Pathological non-IID}
        \label{fig:sun397_pat_fs}
    \end{subfigure} \hspace{0em}%

    \vspace{-1.0em}

    \begin{subfigure}[b]{0.492\linewidth}
        \includegraphics[width=\linewidth]{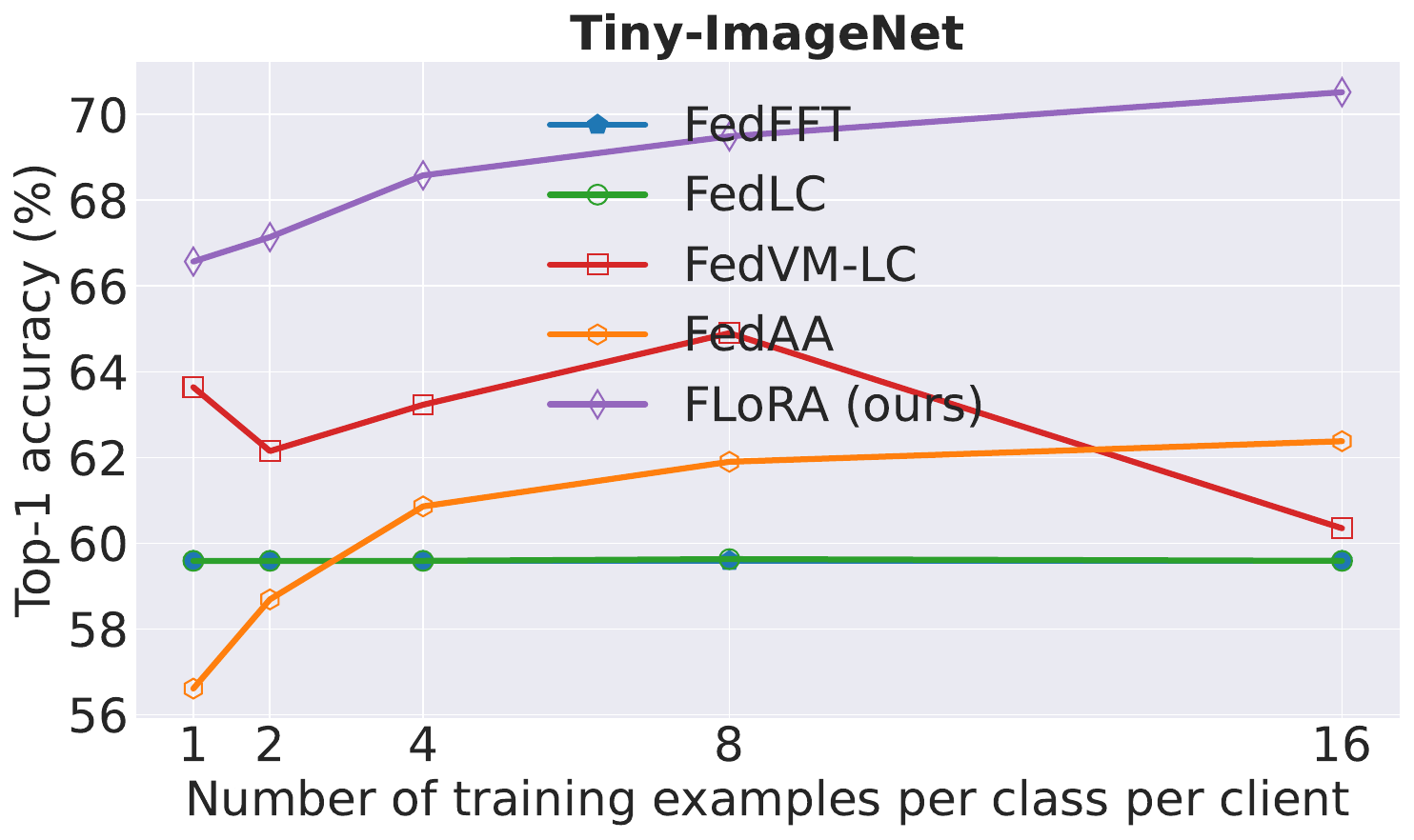}
        \caption{IID}
        \label{fig:tiny_iid_fs}
    \end{subfigure} \hspace{0em}%
    \hfill
    \begin{subfigure}[b]{0.492\linewidth}
        \includegraphics[width=\linewidth]{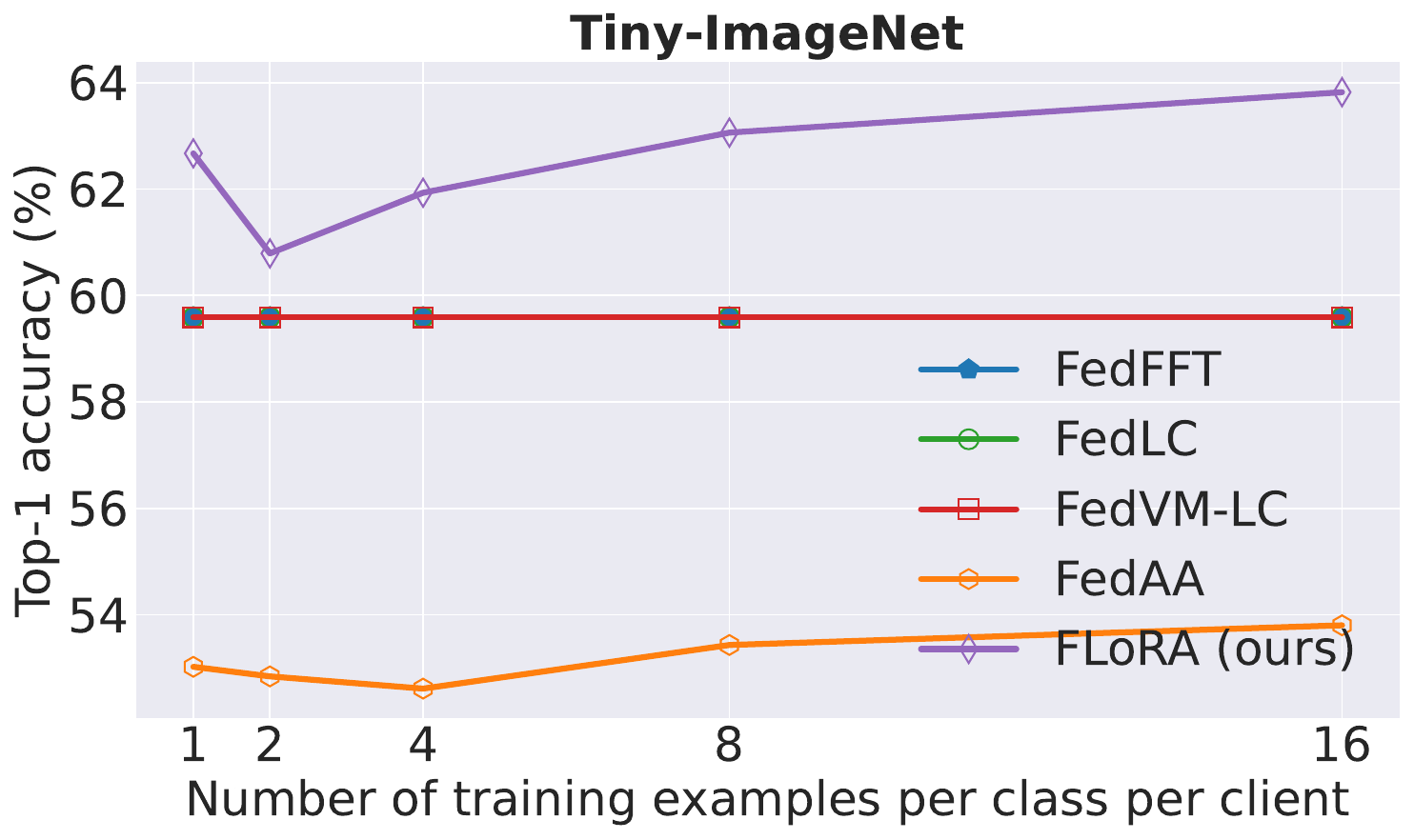}
        \caption{Pathological non-IID}
        \label{fig:tiny_pat_fs}
    \end{subfigure} \hspace{0em}%

    \caption{Result of few-shot learning on selected datasets.}
    \label{fig:fs}
\end{center}

\vspace{-2.0em} 

\end{figure}

\subsection{Efficiency Analysis}

In \proj, a key attribute is that we incorporate a LoRA adapter independent of the datasets. Hence, it only communicates the tiny size network through training and inference, resulting in lower communication costs than FL algorithms that use whole models for aggregation. We evaluate the communication cost in way such that we train all the models to a target accuracy and calculate the communication cost. The communication cost is represented by:
\begin{equation}
\label{eq:cost}
    N \times \rho \times
    \text{cost per round}, 
\end{equation}
where the cost per round is $2 \times$ the size of the exchanged model (for downloading and uploading the transferred model). The transferred model size details are given in Table~\ref{tab:communication}.

As demonstrated in Figure~\ref{fig:overhead}, our method, \proj, achieves the target accuracies in fewer communication rounds while substantially reducing both the communication and training costs, as analyzed for the CIFAR-10, Flowers102, and FER2013 datasets. Notably, in the IID setting for CIFAR-10, \proj requires just one communication round—significantly fewer than the 15 rounds required by competing methods like FedLC. This efficiency carries over to non-IID settings, where \proj also outperforms in communication costs and training time, suggesting rapid convergence capabilities and suitability for practical federated learning deployments.

Figure~\ref{fig:memory_all} further reveals that \proj maintains a lower GPU memory usage of $11.84$ GB, offering an optimal balance between resource economy and model performance. This is attributed to the strategic integration of LoRA adapters, which refine a small yet impactful subset of model parameters, enhancing computational efficiency—a crucial advantage for deploying advanced machine learning on devices with constrained resources.

The efficiency gains from \proj suggest the model's potential to facilitate federated learning in environments where minimizing resource use is as important as maintaining data privacy and reducing bandwidth. These findings underscore the viability of \proj as an effective solution to the challenges of scalability and communication overhead in distributed learning contexts.

\subsection{Few-Shot Evaluation}

Few-shot learning is a challenging machine learning paradigm where models must learn from a very limited amount of data. This is particularly relevant in FL, where each client may only have a small number of examples.

In addition to IID scenario, we create pathological heterogeneous setting to simulate the FL environment~\cite{mcmahan2017fedavg,shamsian2021pfedfn}.
For the pathological label skew setting, we sample data with a fraction of total label amount for each client on each dataset from their total of categories, with disjoint data and different numbers of data samples. The details of the number of classes per client in the pathological setting are presented in Table~\ref{tab:fs_classes_vertical}.

\subsubsection{Homogeneous setting}

Each client will have $n-$shot training samples of each class (each client has every class). There is a common testing set by randomly selecting $20\%$ of the total number of samples. The data for testing is selected from the remaining samples after allocating the training data, ensuring there is no overlap between training and testing sets. The experiment will be carried up to $16-$ shot. Due to the limited total number of samples of serveral datasets, we are only able to do as few as $4-$ shot learning.

\subsubsection{Pathological heterogeneous setting}

Here, each client owns $k$ disjoint classes with n-shot training samples of each of these $k$ classes if the number of classes is divisible by the number of clients ($N$) to evenly distribute $k$ classes per client. For the cases where the number of classes is not perfectly divisible by the number of clients, we distribute the remainder of the classes to the last client. The experiment will be carried up to $16-$ shot.

In the analysis of few-shot learning, \proj emerges as a strong method in federated learning contexts. It outshines established baselines across $16$ datasets, exhibiting a robust capability to generalize well and optimize performance even with limited data availability, as depicted in Figure~\ref{fig:fs_avg}. The empirical data show that \proj not only thrives in IID settings, consistently delivering high top-1 accuracy as the number of per-client training examples increases, but it also maintains a competitive edge in pathological settings that mimic real-world non-IID conditions.

Despite the inherent challenges presented by few-shot learning, \proj's performance suggests a particular proficiency in managing data scarcity and distribution variability. While other methods like FedFFT and FedVM-LC plateau or falter, especially when faced with highly skewed data, \proj's consistent accuracy—even with minimal training examples—underscores its potential in practical applications that must contend with diverse and imbalanced datasets. Figure~\ref{fig:fs} shows the results of two selected datasets with relatviely big number of classes, SUN397 and TINY. These insights reinforce the relevance of \proj for few-shot federated learning and underscore its potential impact on the broader deployment of machine learning solutions in data-constrained environments.

\section{Ablations}


\begin{table*}[ht]
\caption{Different configurations of LoRA within CLIP. We have selected $r=2$ because it offers a good tradeoff between accuracy and number of trainable parameters. The scalaing factor $\alpha$ does not affect the number of trainable parameters.}
\label{tab:ablation_size}
\begin{center}
\begin{tabular}{c|cccccc|cccccc}
\toprule
& \multicolumn{6}{|c}{\textbf{Text Encoder}} & \multicolumn{6}{|c}{\textbf{Image Encoder}} \\ 
\midrule
\multicolumn{1}{c|}{\textbf{Rank}} & r = 1 & \cellcolor{blue!25} r = 2 & r = 4 & r = 8 & r = 16 & r = 32 & r = 1 & r = 2 & r = 4 & r = 8 & r = 16 & r = 32 \\ 
\midrule
\textbf{LoRA size}    & 12,288 & 24,576 & 49,152 & 98,304 & 196,608 & 393,216 & 18,423 & 36,864 & 73,728 & 147,456 & 294,912 & 589,824 \\
\bottomrule
\end{tabular}
\end{center}

\vspace{-1.0em}

\end{table*}


\begin{figure}[ht]

\begin{center}
    \begin{subfigure}[b]{0.492\linewidth}
        \includegraphics[width=\linewidth]{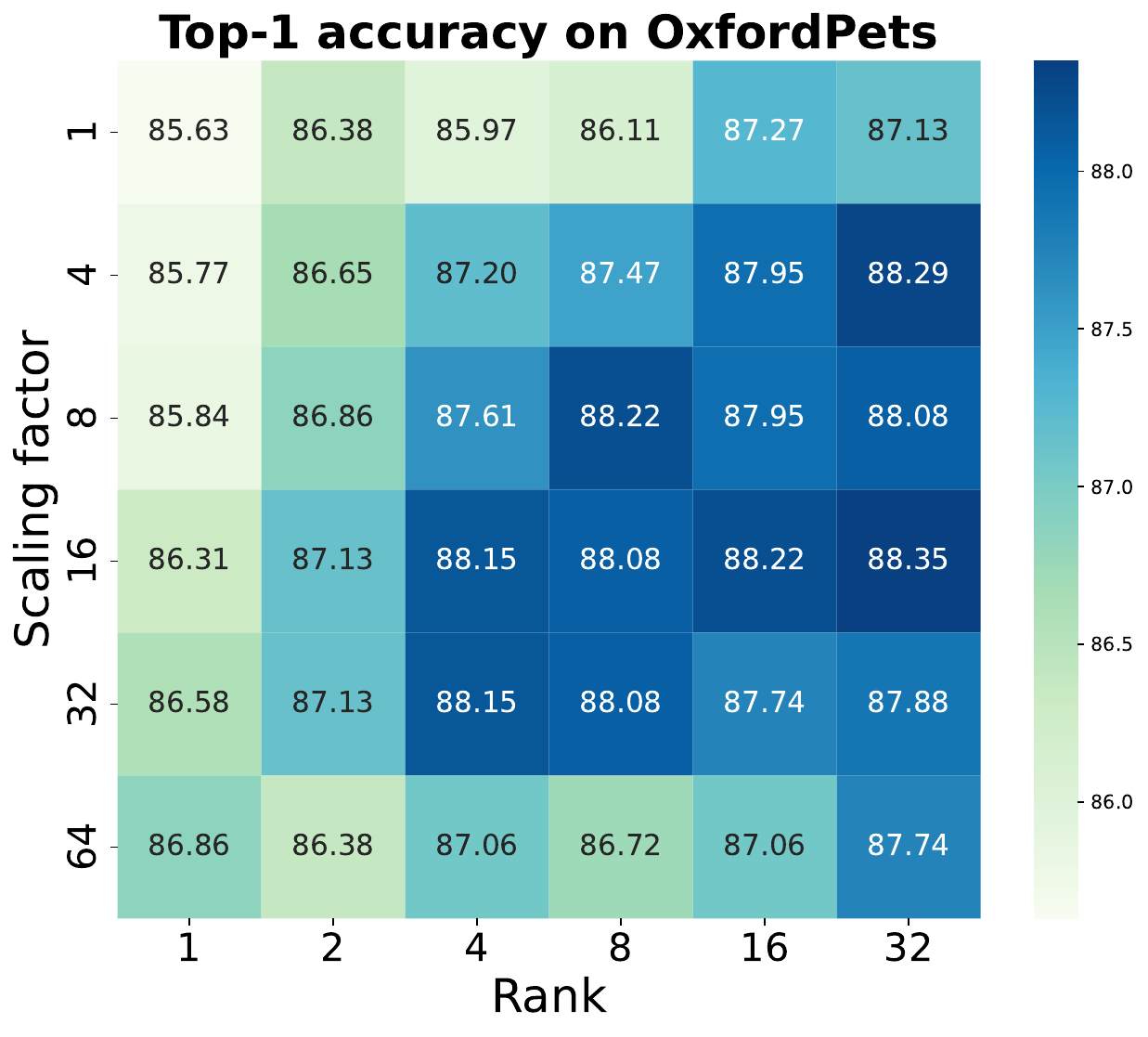}
        \label{fig:p37_projection_text}
    \end{subfigure} \hspace{0em}%
    \hfill
    \begin{subfigure}[b]{0.492\linewidth}
        \includegraphics[width=\linewidth]{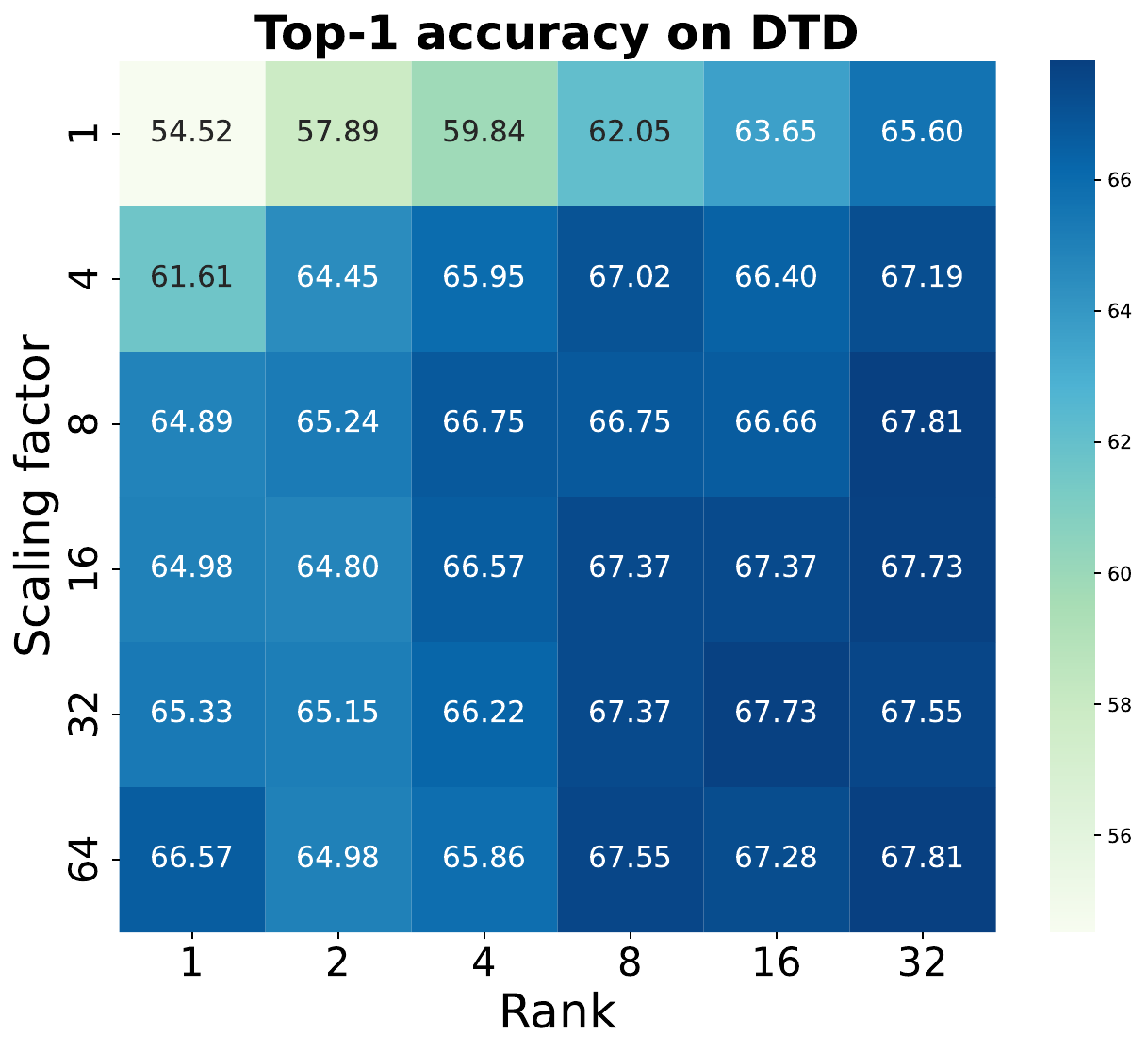}
        \label{fig:d47_projection_text}
    \end{subfigure} \hspace{0em}%

    \vspace{-2.0em}

    \begin{subfigure}[b]{0.492\linewidth}
        \includegraphics[width=\linewidth]{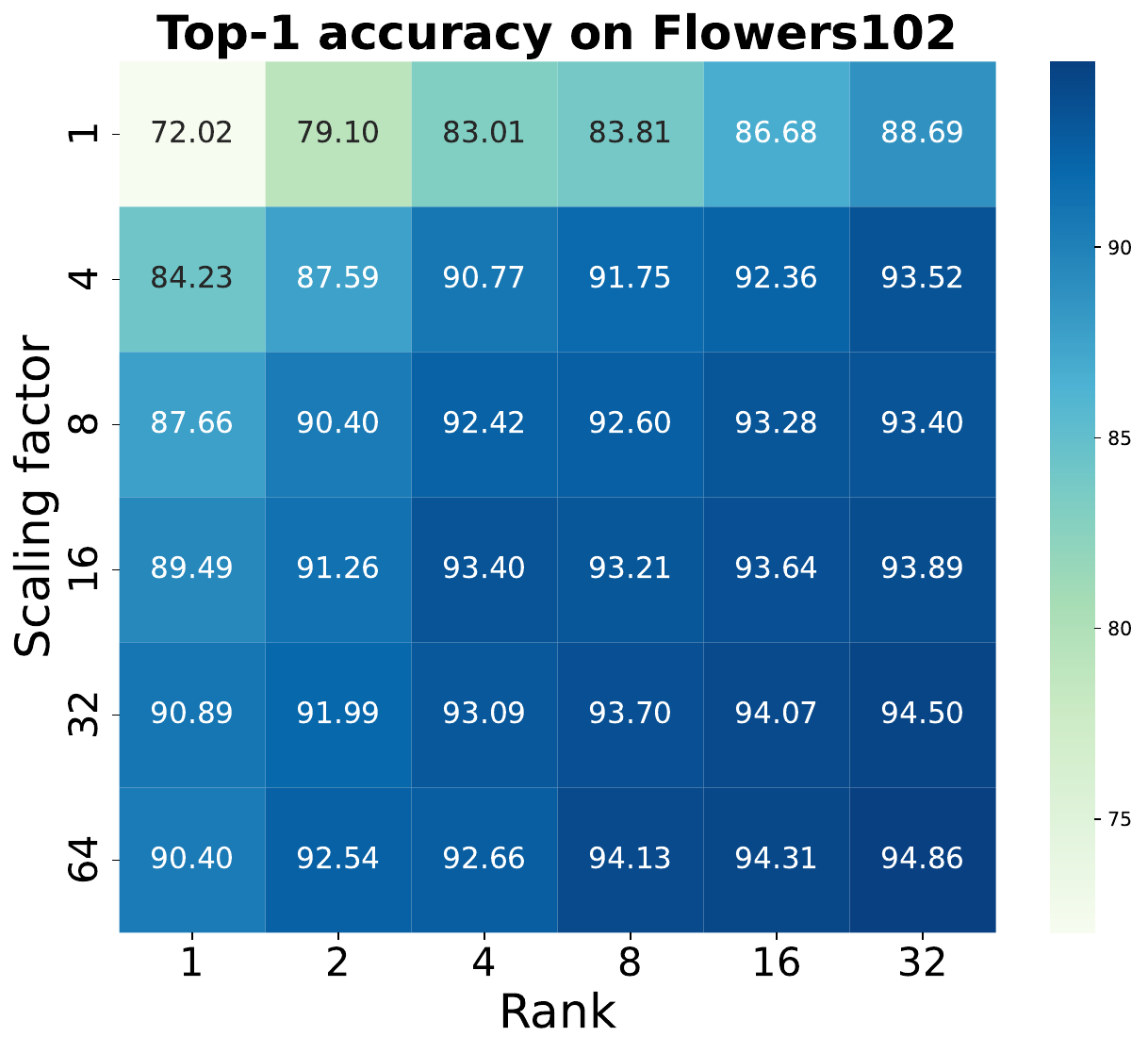}
        \label{fig:f102_projection_text}
    \end{subfigure} \hspace{0em}%
    \hfill
    \begin{subfigure}[b]{0.492\linewidth}
        \includegraphics[width=\linewidth]{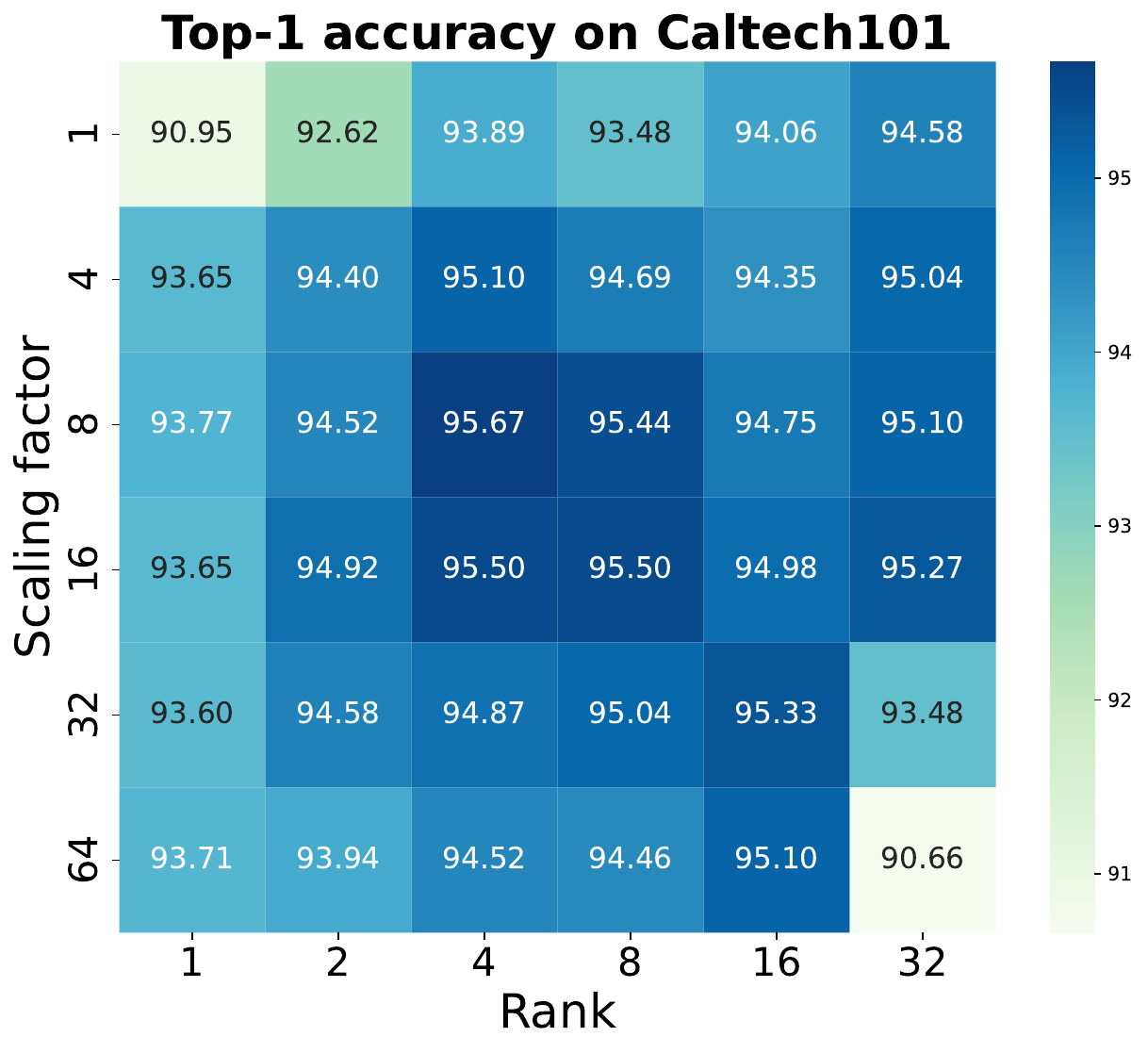}
        \label{fig:c101_projection_text}
    \end{subfigure} \hspace{0em}%

    \caption{Different configurations of LoRA adapters placed on projection layers on CLIP's text encoder in non-IID setting ($\beta=0.1$).}
    \label{fig:dir_projection_text}
\end{center}

\vspace{-2.0em}

\end{figure}

\begin{figure}[ht]

\begin{center}
    \begin{subfigure}[b]{0.492\linewidth}
        \includegraphics[width=\linewidth]{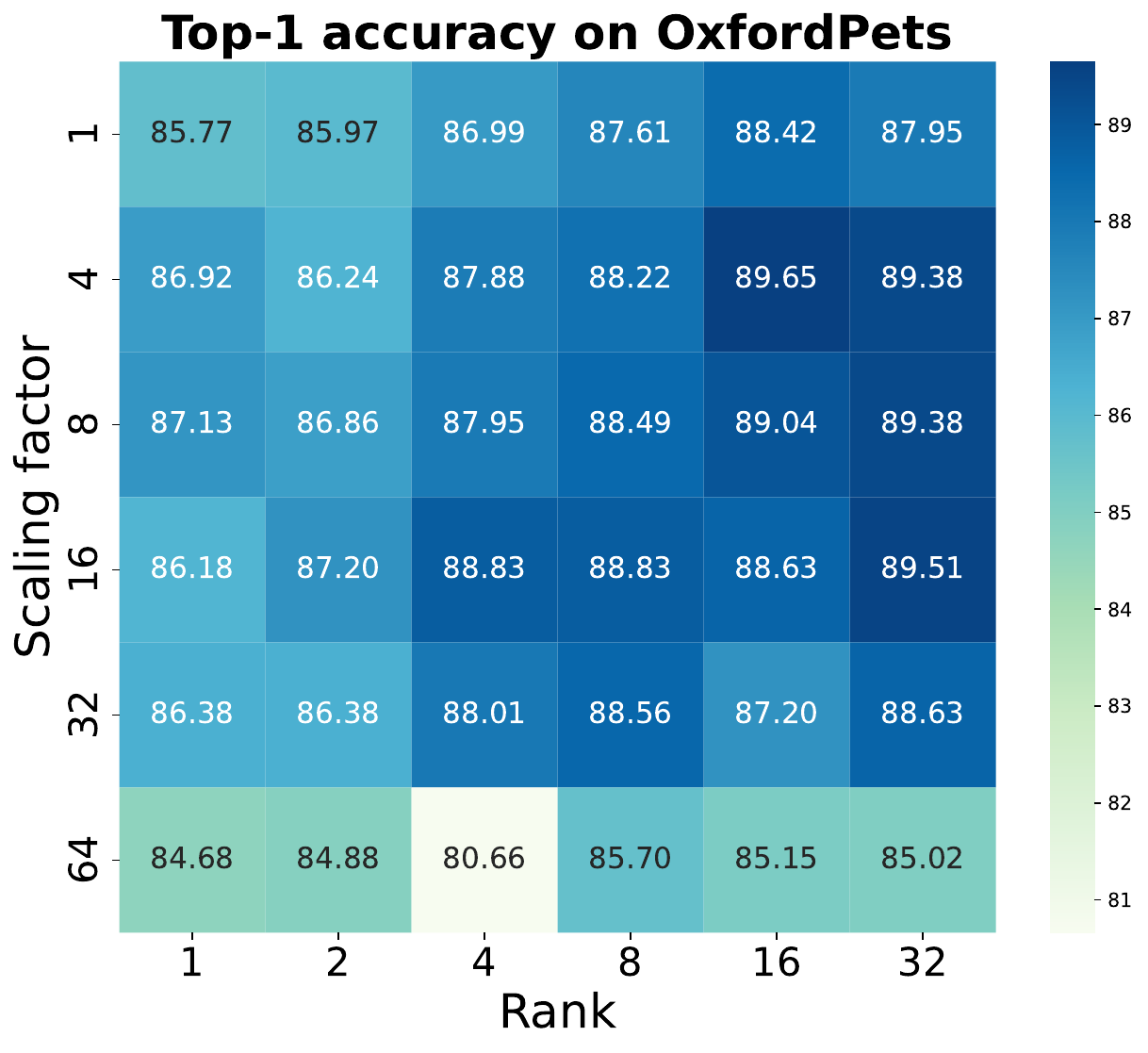}
        \label{fig:p37_projection_vision}
    \end{subfigure} \hspace{0em}%
    \hfill
    \begin{subfigure}[b]{0.492\linewidth}
        \includegraphics[width=\linewidth]{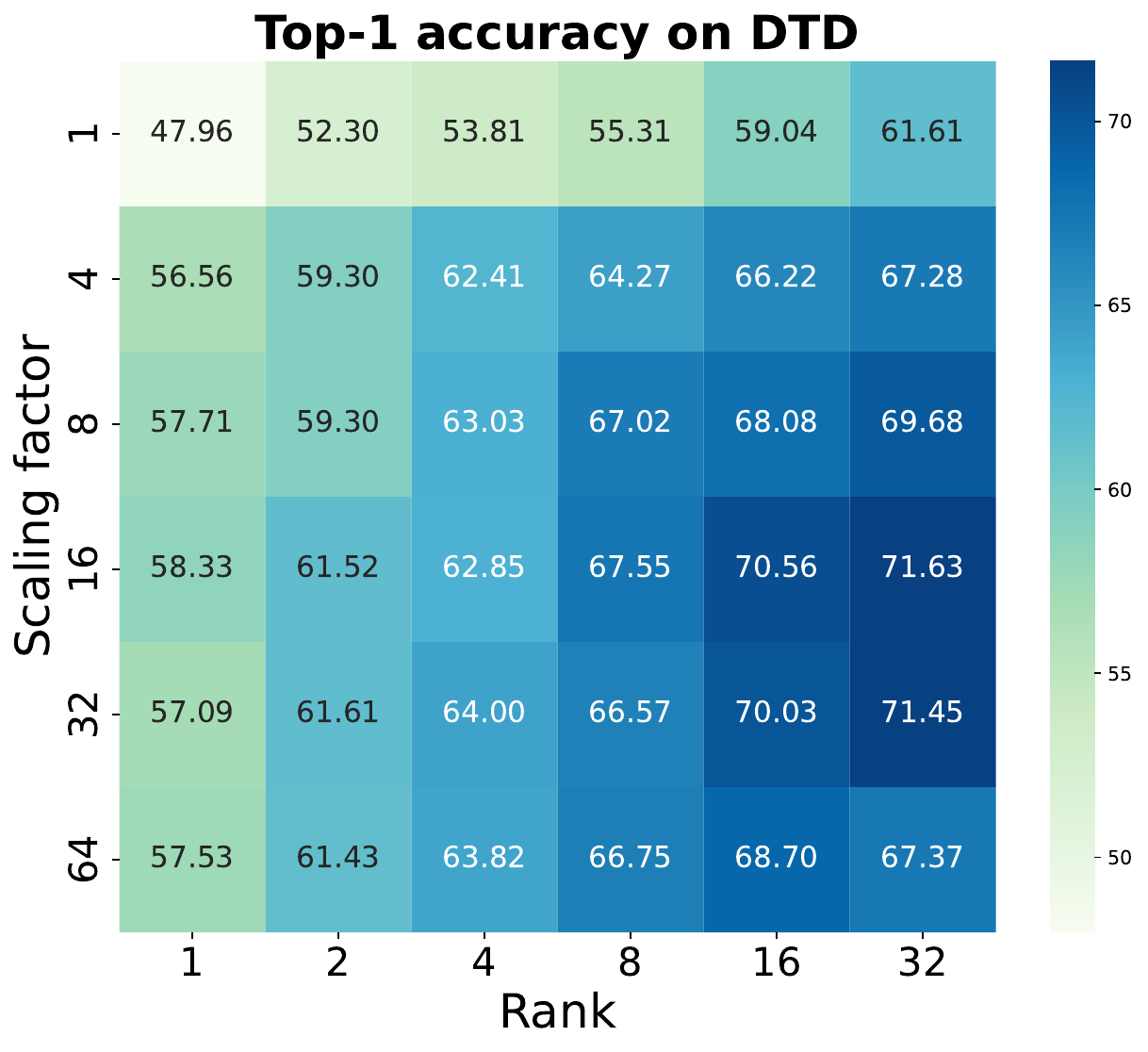}
        \label{fig:d47_projection_vision}
    \end{subfigure} \hspace{0em}%

    \vspace{-2.0em}

    \begin{subfigure}[b]{0.492\linewidth}
        \includegraphics[width=\linewidth]{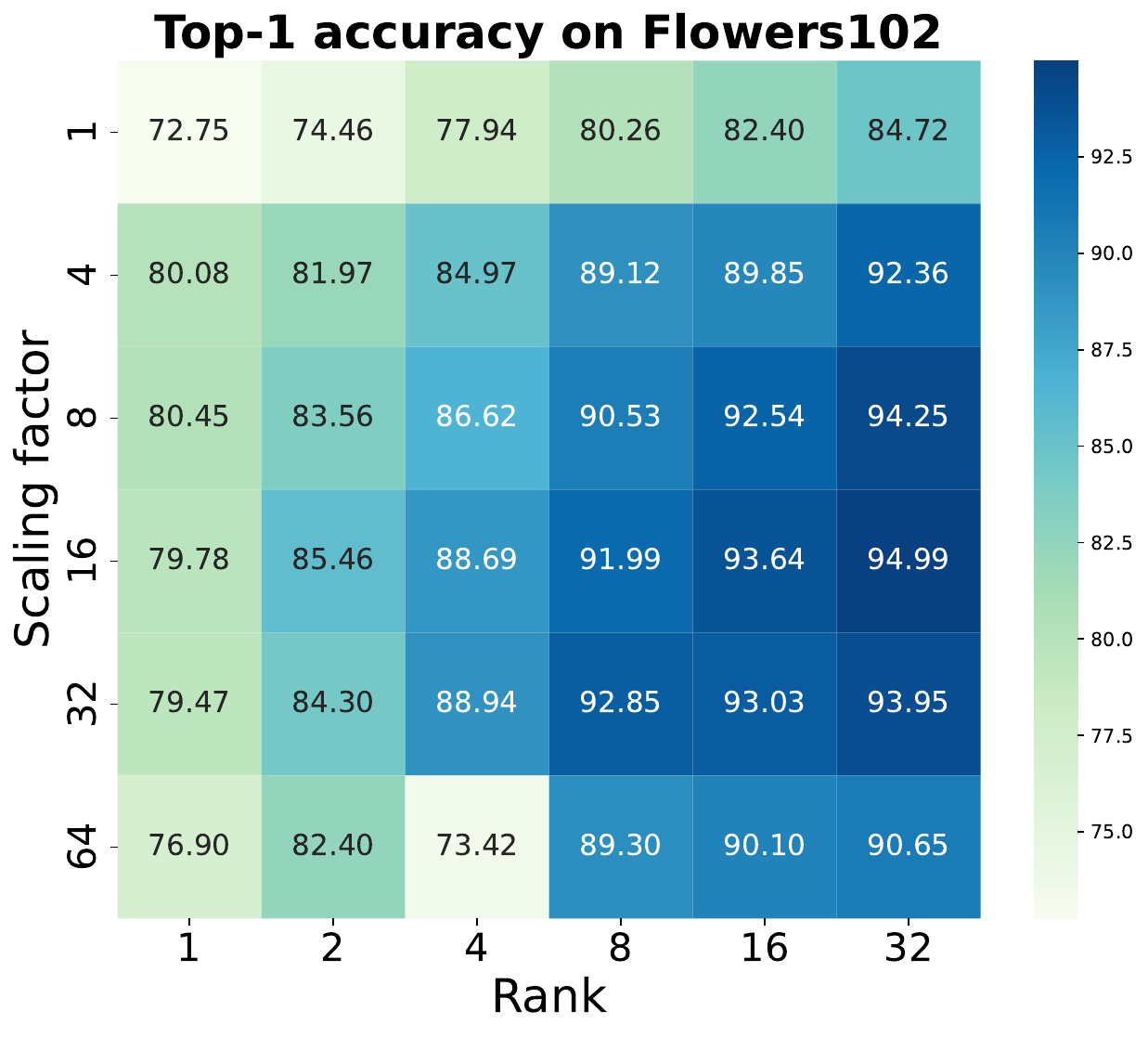}
        \label{fig:f102_projection_vision}
    \end{subfigure} \hspace{0em}%
    \hfill
    \begin{subfigure}[b]{0.492\linewidth}
        \includegraphics[width=\linewidth]{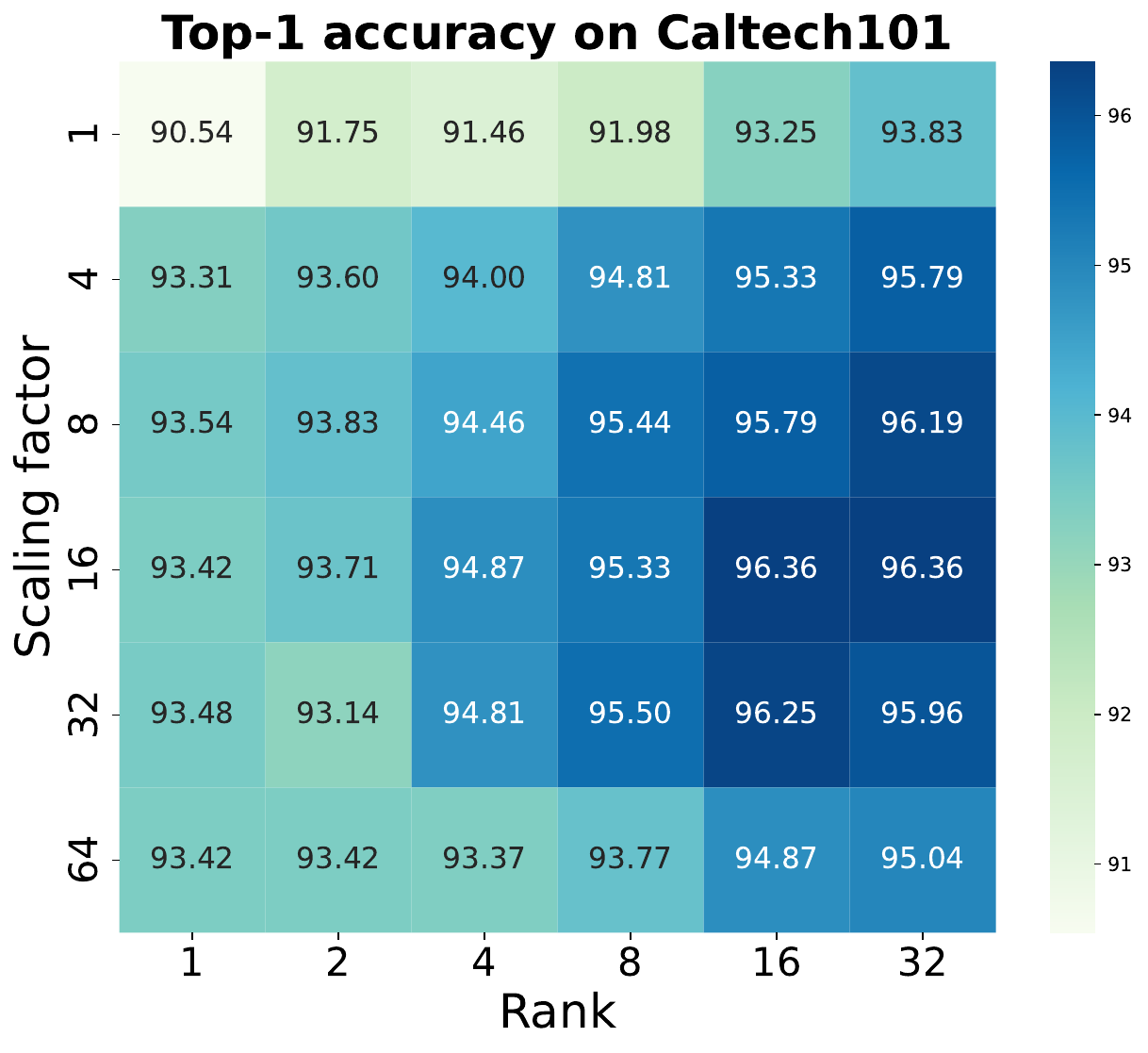}
        \label{fig:c101_projection_vision}
    \end{subfigure} \hspace{0em}%

    \caption{Different configurations of LoRA adapters placed on projection layers on CLIP's Image Encoder in non-IID setting ($\beta=0.1$).}
    \label{fig:dir_projection_vision}
\end{center}

\vspace{-2.0em}

\end{figure}

The ablation study we conducted rigorously evaluates the impact of varying LoRA hyperparameters—scaling factors and ranks—on the performance of the CLIP model's text and image encoders within a non-IID federated learning setting. As delineated in Figures~\ref{fig:dir_projection_text} and ~\ref{fig:dir_projection_vision}, the study dissects the individual and synergistic effects of these hyperparameters on the top-1 accuracy across four distinct datasets, with a particular focus on a non-IID setting characterized by $\beta = 0.1$.

The study explores a range of scaling factors $(1, 4, 8, 16, 32, 64)$ and ranks $(1, 2, 4, 8, 16, 32)$ to understand their individual and combined impact on model performance. The scaling factor likely pertains to the amplification of the projection layer's adjustments, while rank could represent the number of modifications within the adapter that are allowed to vary.

Our findings, summarized in Table~\ref{tab:ablation_size}, reveal that a LoRA adapter with a rank of $2$ strikes a strategic balance, significantly reducing the number of trainable parameters to $24,576$ for the text encoder and to $36,864$ for the image encoder, while preserving competitive accuracies. This balance showcases the potential for achieving considerable model performance with a reduced computational footprint—a key factor in the resource-constrained environments typical of federated learning.

Notably, the integration of the LoRA adapter into the text encoder consistently outperforms the image encoder integration while simultaneously maintaining a lower parameter count. This indicates a clear preference for text encoder adaptation in scenarios where both performance optimization and resource efficiency are desired.

The conclusive evidence from our study attests to the rank $2$ LoRA adapter's role in bolstering federated learning efficiency. By adopting this configuration, FedLoRA presents as an optimized framework that promises not only swift model convergence and communication efficacy but also practical applicability across various federated learning contexts, especially those with stringent resource limitations.

Our ablation study examined a series of scaling factors, finding that a judicious selection of $\alpha$ is essential for achieving the best trade-off between model accuracy and computational efficiency. While a larger $\alpha$ may potentially lead to greater improvements in accuracy by allowing more pronounced updates to the model, it also risks overfitting and straying too far from the valuable features learned during pre-training. Conversely, a smaller $\alpha$ ensures that the updates remain subtle, preserving the pre-trained model's generalizability but may not capture the nuances of the new task as effectively.

The empirical results from our study suggest that a moderate scaling factor offers the most effective balance, allowing for sufficient adaptation to the target task without overwhelming the pre-existing strengths of the model. By carefully calibrating $\alpha$, we enable FedLoRA to adapt to new data while maintaining the integrity of the original CLIP architecture, thus ensuring that our approach is not only resource-conscious but also retains the rich representational capabilities of the foundational model.

Incorporating LoRA within the CLIP model presents a wealth of potential beyond the scope of this paper, such as adapting LoRA to other components like the query, value, key, and multi-layer perceptron elements, as well as across different heads within the model's architecture. Each of these elements offers a unique set of possibilities for parameter-efficient adaptation with varying ranks and scaling factors. While our current study does not extend to these configurations, they represent intriguing avenues for future investigation. Exploring these alternatives could further illuminate the ways in which federated learning can benefit from LoRA's efficient fine-tuning, potentially leading to even more sophisticated and resource-aware federated learning models.






\section{Conclusion}




In this paper, we have rigorously evaluated the efficacy of federated learning (FL) techniques for fine-tuning the CLIP model via LoRA adapters across a spectrum of IID, non-IID, and few-shot learning settings. Our proposed methodology, \proj, consistently surpasses conventional FL benchmarks, demonstrating superior accuracy and adaptability to diverse datasets and learning environments.

The significant performance gains of \proj are highlighted in our extensive evaluations, which reveal its remarkable competence in handling both data abundance and scarcity. Ablation studies further distill the critical hyperparameters that underpin the success of multimodal FL models, ensuring both adaptability and computational efficiency.

Ultimately, our investigation into \proj advances the field of FL by showcasing a method that not only enhances model performance in distributed settings but also offers a strategic blueprint for efficient multimodal model fine-tuning. The results of our work suggest promising directions for future FL systems poised to navigate the complexities of practical data challenges.

In conclusion, this paper presents a step forward in FL, proposing a method that improves the accuracy of models trained across decentralized data and introduces a framework for efficiently fine-tuning multimodal models in a federated context. Our findings open up new avenues for research into adaptive FL systems, which can handle real-world data intricacies while maintaining user privacy and minimizing communication costs.

\bibliographystyle{IEEEtran}
\bibliography{ref}

\clearpage

\clearpage



















\clearpage

\end{document}